%% file: main.tex
\documentclass[]{ztxtech}
\hidelogo

\usepackage{microtype}
\usepackage{graphicx}
\usepackage{subcaption}
\usepackage{booktabs}
\usepackage{multirow}
\usepackage{rotating}
\usepackage{colortbl}
\usepackage{xcolor}
\usepackage{array}
\usepackage{ifthen}
\usepackage{adjustbox}
\usepackage[para,online,flushleft]{threeparttable}
\usepackage{bbding}
\usepackage{makecell}
\usepackage{nicefrac}
\usepackage{amsfonts}
\usepackage{url}
\usepackage{xspace}
\usepackage{fontawesome5}
\usepackage{titletoc}

\usepackage{amsmath}
\usepackage{amssymb}
\usepackage{mathtools}
\usepackage{amsthm}
\usepackage[ruled, vlined]{algorithm2e}
\usepackage[normalem]{ulem}

\colorlet{ztxtechhl}{ztxtechpurple!18!white}
\newtcolorbox{painpointbox}{
    enhanced,
    breakable,
    colback=ztxtechpurple!6!white,
    colframe=ztxtechpurple!18!white,
    boxrule=0.45pt,
    arc=8pt,
    left=7pt,
    right=7pt,
    top=8pt,
    bottom=8pt,
    before skip=8pt,
    after skip=8pt
}
\newtcbox{\hl}{on line,
    enhanced,
    box align=base,
    colback=ztxtechhl,
    colframe=ztxtechhl,
    boxrule=0pt,
    arc=4pt,
    left=2pt,
    right=2pt,
    top=1pt,
    bottom=1pt,
    boxsep=0pt,
    nobeforeafter}

\newtcbox{\dbhl}{on line,
    enhanced,
    box align=base,
    colback=gray!80,
    colframe=gray!80,
    boxrule=0pt,
    arc=4pt,
    left=2pt,
    right=2pt,
    top=1pt,
    bottom=1pt,
    boxsep=0pt,
    nobeforeafter,
    coltext=white}

\newcommand{\method}{\hl{\textbf{\texttt{Aion}} \methodlogo}\xspace}
\newcommand{\directbuild}{\dbhl{\texttt{Direct Build}}\xspace}
\newcommand{\repo}{https://github.com/ztxtech/aion}
\AtBeginDocument{%
    \pdfstringdefDisableCommands{%
        \def\method{Aion}%
        \def\hl#1{#1}%
        \def\dbhl#1{#1}%
    }%
}

\theoremstyle{plain}
\newtheorem{theorem}{Theorem}[section]

\theoremstyle{definition}
\newtheorem{definition}[theorem]{Definition}

\theoremstyle{remark}

\title{\method: Next-Generation Tasks and Practical Harness for Time Series}

\author[1]{Tianxiang Zhan}
\author[2,1]{Xiaobao Song}
\author[3,1]{Tong Guan}
\author[1]{Shirui Pan}
\author[1]{Ming Jin}

\affiliation[1]{Griffith University}
\affiliation[2]{Shenzhen University}
\affiliation[3]{Zhejiang University}

\abstract{
Time series research is moving beyond fixed forecasting benchmarks toward realistic tasks that combine prediction, contextual reasoning, tool use, and structured decision support. Most benchmarks are built around clean data and short evaluation loops; agents alone may miss temporal constraints, evidence checks, or review before finalizing outputs. We first formalize next-generation time series tasks as three-component tuples consisting of a task file, a workspace, and a validation interface. We then present \method, a time series harness built from six component groups: agents, skills, rules, memory, evaluation, and protocols. In this harness, we use three design principles: temporal grounding, temporal knowledge-grounded reasoning, and reliability mechanisms such as post-experiment analysis and layered review. A Kaggle Store Sales case study shows that the harness produces more detailed process traces, more artifacts, and more review steps than the same base agent operating in OpenCode \directbuild mode. Taken together, these results argue for a paradigm shift from fixed tasks to realistic ones under real-world constraints.
}

\date{\today}
\correspondence{Shirui Pan \email{s.pan@griffith.edu.au}; Ming Jin \email{ming.jin@griffith.edu.au}}
\ztxtechdata[Code]{\url{\repo}}

\begin{document}

\maketitle

\section{Introduction}\label{sec:introduction}

\begin{figure}[h]
    \centering
    \includegraphics[width=\textwidth]{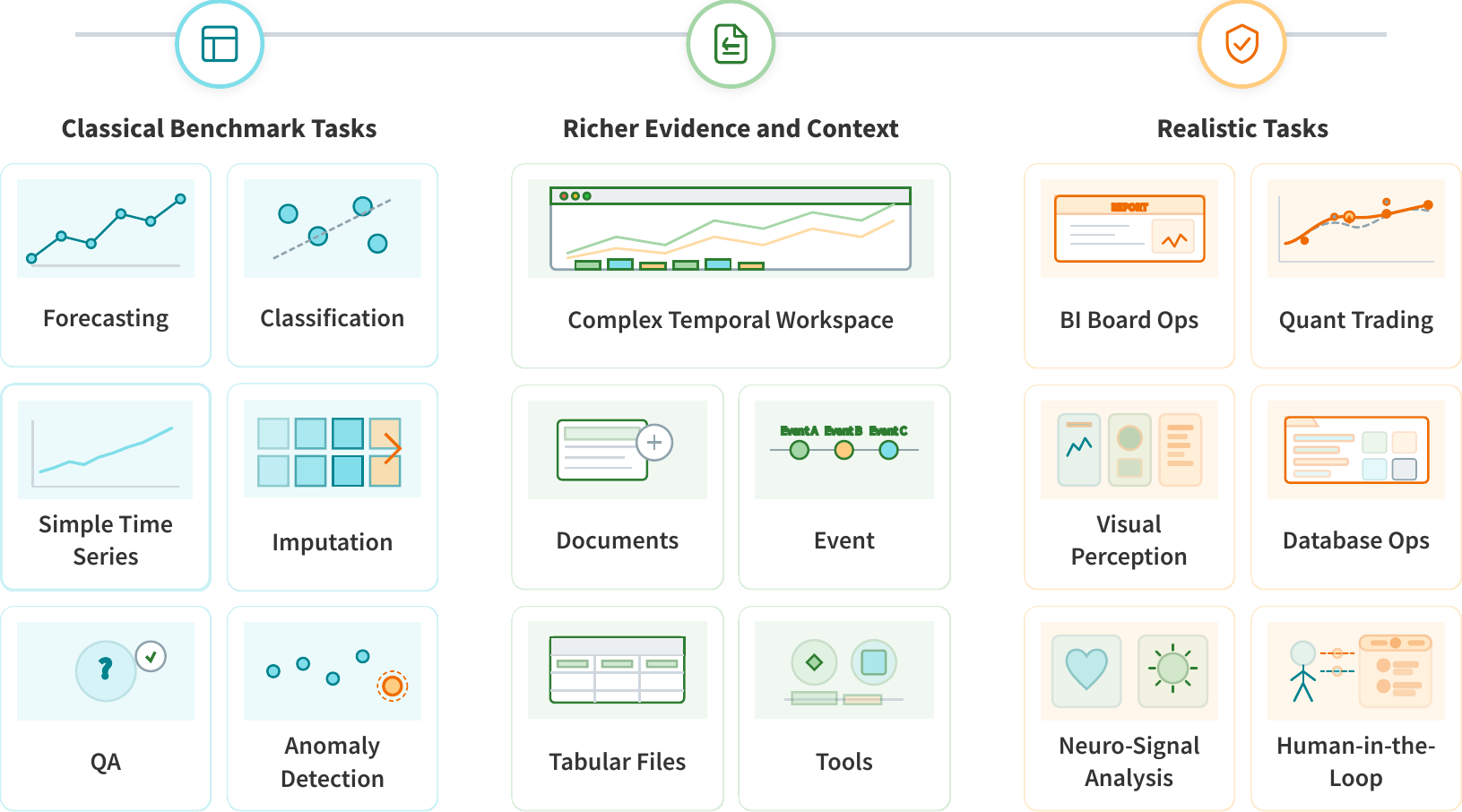}
    \caption{The shift from analytical tasks to realistic time series problem-solving with broader evidence and complex workspaces.}
    \label{fig:intro-workload-shift}
\end{figure}

Time series research has expanded well beyond standalone forecasting models~\citep{jin2023large,goswami2024moment,shao2024exploring}. Recent progress spans large benchmark toolkits, foundation models~\citep{liang2024foundation,qiu2024tfb}, multimodal forecasting settings~\citep{chen2025mtbench}, temporal reasoning benchmarks~\citep{ye2024domain, yin2026mmts}, and agentic systems for analysis and prediction \citep{liu2025moirai,liu2025ts,weng2026temporalbench}. The problem itself is also changing. Realistic time series tasks are no longer limited to extrapolating future values from historical signals. They increasingly require reasoning over context~\citep{kong2025time}, events~\citep{ge2025eventtsf}, interventions, domain constraints, and downstream decisions across domains such as retail~\citep{makridakis2020m5}, healthcare~\citep{johnson2023mimic}, transportation~\citep{STGMMOE}, and physical systems~\citep{kaur2023codit}. In many realistic settings~\citep{liu2023agentbench}, the system is not asked to output only a forecast~\citep{cai2025timeseriesgym}. For example, it may need to satisfy submission rules on a competition platform~\citep{chan2024mle}, produce a structured analytical report, justify a temporal argument from multimodal context, or support a multi-step operational decision. The practical unit of work is therefore no longer a standalone task evaluated with limited metrics. Figure~\ref{fig:intro-workload-shift} summarizes this shift from performing classical analytical tasks to realistic time series problem-solving.

Current time series benchmarks are still built around a relatively fixed set of problems~\citep{wang2024deep}, mainly about forecasting, classification~\citep{herzen2022darts}, imputation~\citep{du2023pypots,wangj2024deep}, anomaly detection~\citep{wenig2022timeeval}, and question answering~\citep{kong2025time}. These tasks are well-established but they usually assume that the problem has already been simplified to clean datasets paired with designated evaluation metrics. That assumption is overly narrow for many real scenarios, where a comprehensive solution should combine temporal understanding~\citep{bommasani2023holistic}, contextual reasoning~\citep{chen2025mtbench}, tool use~\citep{zhou2023webarena}, and multi-step decision-making~\citep{mialon2023gaia}. At the same time, large language models (LLMs) and agentic systems are becoming capable of much more complex work in other domains. However, such a capable agent is not automatically a ready-to-use system for time series problem-solving. It may search broadly while violating temporal constraints, such as using the wrong forecast horizon or leaking future information. It may also produce an answer without passing evidence checks or terminate before the relevant checks have been completed. What is missing between benchmarks and existing agent systems is an explicit control framework called a \hl{harness} for how an agent system takes a task, invokes models and tools, tracks intermediate state, enforces constraints, and evaluates outputs~\citep{pan2026natural,anthropic_harnesses,openai_harness}. A harness orchestrates task interpretation, tool use, validation, and review in one execution loop. Its role is to connect task specification, workspace exposure, runtime control, validation, and evaluation into one execution process.
We need a \textbf{time series harness} that coordinates the solving process, tracks intermediate evidence, and enforces validation before delivery.
This motivates the two critical questions below.

\begin{itemize}
    \item \textbf{Q1:} What problem sets should the \textbf{next-generation time series intelligent systems} focus on?
    \item \textbf{Q2:} How can agents serve as \textbf{reliable components} for solving real-world time series tasks?
\end{itemize}

We address these gaps in two steps. First, we define the research problem as solving \textbf{next-generation time series tasks} with task files, workspaces, and validation interfaces. This answers \textbf{Q1} and keeps practical complications such as missing values, partial column exposure, irregular timing, nonstandard utilities, and mixed deliverables inside the problem. Second, we present \method as a \textbf{time series harness} that answers \textbf{Q2}. \method is built from six component groups in Section \ref{sec:method-overview}. Under this design, we use three design principles: \emph{temporal grounding, temporal knowledge-grounded reasoning, and reliability mechanisms.} We evaluate the framework through a Kaggle case study. The study reconstructs the Store Sales competition as a local forecasting task and compares the proposed harness run with the same base model in OpenCode's direct build mode. The harnessed run reaches a lower final RMSLE while adding review blocks, critique rounds, and figures to the execution trace. This contrast suggests that the proposed harness not only improves the scored submission but also keeps strategy checking and completion review. In short, our main contributions are as follows.

\begin{itemize}
    \item We introduce \textbf{next-generation time series tasks} defined by task files, workspaces, and validation interfaces. This shifts the field from benchmarking against isolated dataset-and-metric pairs toward solving structured, constraint-aware problem formulations.
    \item We present \method as a \textbf{time series harness} built from \emph{agents}, \emph{skills}, \emph{rules}, \emph{memory}, \emph{evaluation}, and \emph{protocols}. We further organize the harness around three time series design principles: temporal grounding, temporal knowledge-grounded reasoning, and reliability mechanisms, which keep temporal constraints, external evidence, and review steps inside the solving process.
\end{itemize}

\section{Related Work}\label{sec:related-work}

\subsection{Toward Realistic Time Series Tasks}\label{sec:classical-tasks-and-domain-preferences-in-time-series}

Classical time series research mainly focuses on several well-established tasks~\citep{wang2024deep,shao2024exploring}, including forecasting~\citep{benidis2022deep}, classification~\citep{herzen2022darts}, imputation~\citep{du2023pypots,wangj2024deep}, and anomaly detection~\citep{wenig2022timeeval}, with clean datasets~\citep{zhou2021informer,makridakis2020m5}, various model families~\citep{jin2023time,ansari2024chronos}, and benchmark toolkits~\citep{aksu2024gift,shao2024exploring} supporting this line of work.
Recent studies have started to move beyond these classical tasks. They increasingly ask whether a system can interpret historical structure, incorporate contextual information, account for event effects, and explain its decision in a form that supports downstream decisions. This shift is especially visible across domains~\citep{kong2025time,liu2024timemmd,guan2026timeomnivlunifiedmodelstime}. Retail and operations tasks~\citep{arsenault2025survey, abolghasemi2024machine} often care about promotions, inventory pressure, and reporting requirements. Healthcare tasks~\citep{morid2023time, zhang2023applied} place more weight on multivariate alignment, uncertainty, intervention effects, and safe interpretation. Energy~\citep{turowski2024generating} and physical systems~\citep{hota2025evaluating} emphasize forecast horizon, regime changes, control constraints, and robustness under distribution shift. These real-world tasks do not invalidate classical tasks, but they show that the practical unit of work is no longer a single forecasting problem. What matters is a realistic task: a combination of prediction, reasoning, validation, and communication under domain and operational constraints~\citep{guan2026timeomni, weng2026temporalbench}.

\subsection{Limitations of Time Series Benchmarks and Agents}\label{sec:asymmetry-between-time-series-workloads-and-existing-tooling}

A harness is an explicit control framework that connects task specification, system execution, and result evaluation into one process~\citep{pan2026natural,anthropic_harnesses,openai_harness}. This role is especially important for realistic time series tasks, where systems must satisfy specific constraints. From the harness perspective, existing benchmarks and agents each address only one step in solving real-world problems~\citep{cheng2026position}. The main asymmetries fall into these groups.

\begin{itemize}
    \item \textbf{Oversimplified task formulation}: Existing benchmarks often present clean data, specific tasks, and fixed evaluation methods. They are overly simplified compared with the realistic tasks mentioned earlier, so they leave open what next-generation time series systems should actually be asked to solve.
    \item \textbf{Process-blind evaluation}: Existing time series tasks typically evaluate model quality by a small set of predictive metrics. They say little about whether the solver respected domain knowledge, handled missing values and irregular timestamps appropriately, or followed the constraints that matter in real-world use cases. Addressing these concerns requires a \textbf{visible solution trace}~\citep{huang2024mlagentbench} that records not only the final output but also the intermediate decisions and checks that produced it.
\end{itemize}

These limitations argue for a harness that governs the solution process end-to-end and keeps the full trace visible.

\section{\method}\label{sec:projname}
To answer \textbf{Q1} and \textbf{Q2} constructively, we first define \textbf{next-generation time series tasks} as a more realistic task formulation. We then introduce the core idea of \method.

\subsection{Task Specification}\label{sec:task-specification-for-next-generation-time-series-problems}

Real-world tasks are often messy: they may contain missing values, arrive at irregular intervals, and require outputs with explanations. These requirements cannot be adequately specified by a dataset and metric pair alone. We therefore treat the task itself as an explicit specification consisting of a task file, a workspace, and a validation interface. The notation is summarized in Table~\ref{tab:task-notation}.

\begin{table}[h]
    \centering
    \caption{Notation used in this section.}
    \label{tab:task-notation}
    \renewcommand{\arraystretch}{1.14}
    \begin{tabular}{ll}
        \toprule
        Symbol                                           & Meaning                        \\
        \midrule
        $\mathcal{T} = (\phi, \mathcal{W}, \mathcal{V})$ & Problem set                    \\
        $\phi$                                           & Task file                      \\
        $\mathcal{W}$                                    & Workspace                      \\
        $\mathcal{V}$                                    & Validation interface           \\
        $\mathbb{T}$                                     & Access scope                   \\
        $\Sigma$                                         & Prior knowledge                \\
        $\mathcal{O}$                                    & Legal output form              \\
        $\mathcal{C}$                                    & Constraints                    \\
        $A$                                              & Performance evaluation metrics \\
        $o$                                              & Candidate output               \\
        $g$                                              & Validity checks                \\
        $s$                                              & Scored metrics                 \\
        \bottomrule
    \end{tabular}
\end{table}

\begin{definition}\label{def:ngts-task}
    A Next-Generation Time Series Task is a tuple in Eq.~\eqref{eq:ngts-task-triple} that contains three components: a task file $\phi$, a workspace $\mathcal{W}$, and a validation interface $\mathcal{V}$. \uline{These three components answer \textbf{Q1}.}

    \begin{equation}\label{eq:ngts-task-triple}
        \mathcal{T} = (\phi, \mathcal{W}, \mathcal{V}),
    \end{equation}

    \hl{\textbf{Task file.}} $\phi$ is the explicit statement of the problem. It declares the access scope $\mathbb{T}$, the prior knowledge $\Sigma$, the output form $\mathcal{O}$, the constraints $\mathcal{C}$, and the performance evaluation metrics $A$ in Eq.~\eqref{eq:phi}. This is where the task specification includes missing values, hidden columns, irregular timestamps, or downstream utilities (e.g., maximizing profits).

    \begin{equation}\label{eq:phi}
        \phi = (\mathbb{T}, \Sigma, \mathcal{O}, \mathcal{C}, A)
    \end{equation}

    \hl{\textbf{Workspace.}} $\mathcal{W}$ is the exposed system environment.
    It may contain one or more time series files, auxiliary tables, textual instructions, metadata, event records, APIs, and callable tools. We keep $\mathcal{W}$ open-ended to cover all resources available during solving, rather than restricting it to a single clean table with fully known columns.

    \hl{\textbf{Validation interface.}} $\mathcal{V}$ maps a candidate output $o$ to
    \begin{equation}\label{eq:ngts-validation-interface}
        (g, s) = \mathcal{V}(o),
    \end{equation}
    where $g$ records validity checks such as timestamp alignment, leakage boundaries, or format compliance, and $s$ records scored metrics such as forecasting error, profit, or report quality. A candidate is admissible only when all validity checks in $g$ pass, and admissible candidates are ranked by $s$.
\end{definition}

\begin{painpointbox}
    \textbf{Degeneration to classical tasks.} Definition~\ref{def:ngts-task} reduces to a classical forecasting-style task under the following conditions.
    \begin{itemize}
        \item the time steps are regular;
        \item the released data already covers the required columns;
        \item $\mathcal{O}$ is just a fixed-length numeric horizon;
        \item $A$ reduces to one predictive metric such as MAE or MSE.
    \end{itemize}
    Outside this degenerate case, difficulty can also come from missing or partial observations, multiple released files, or nonstandard output requirements. The classic task is regarded as a \textbf{minimal step executed by \method}.
\end{painpointbox}

\subsection{Overview}\label{sec:method-overview}

\method is built from six component groups: \emph{agents}, \emph{skills}, \emph{rules}, \emph{memory}, \emph{evaluation}, and \emph{protocols}. \emph{Agents} are the front-end roles that drive interpretation, evidence collection, solution construction, and review. \emph{Skills} provide reusable priors and structured workflows that stabilize execution across the lifecycle. \emph{Rules} define the admissible boundary of the workflow. \emph{Memory} records progress, decisions, and structural state so that long executions do not drift. \emph{Evaluation} turns runs into explicit assessment through suites, graders, and final acceptance checks. \emph{Protocols} specify how work, review, rechecking, stop signals, and memory updates communicate between different sessions. \uline{Together, these six groups constitute the harness that answers \textbf{Q2}.} In our implementation, each group contains several named components, which are expanded in Appendix~\ref{sec:appendix-components}.

\subsection{Design Principles of Time Series}\label{sec:core-time-series-principle-principles}

\begin{figure}[h]
    \centering
    \includegraphics[width=1\textwidth]{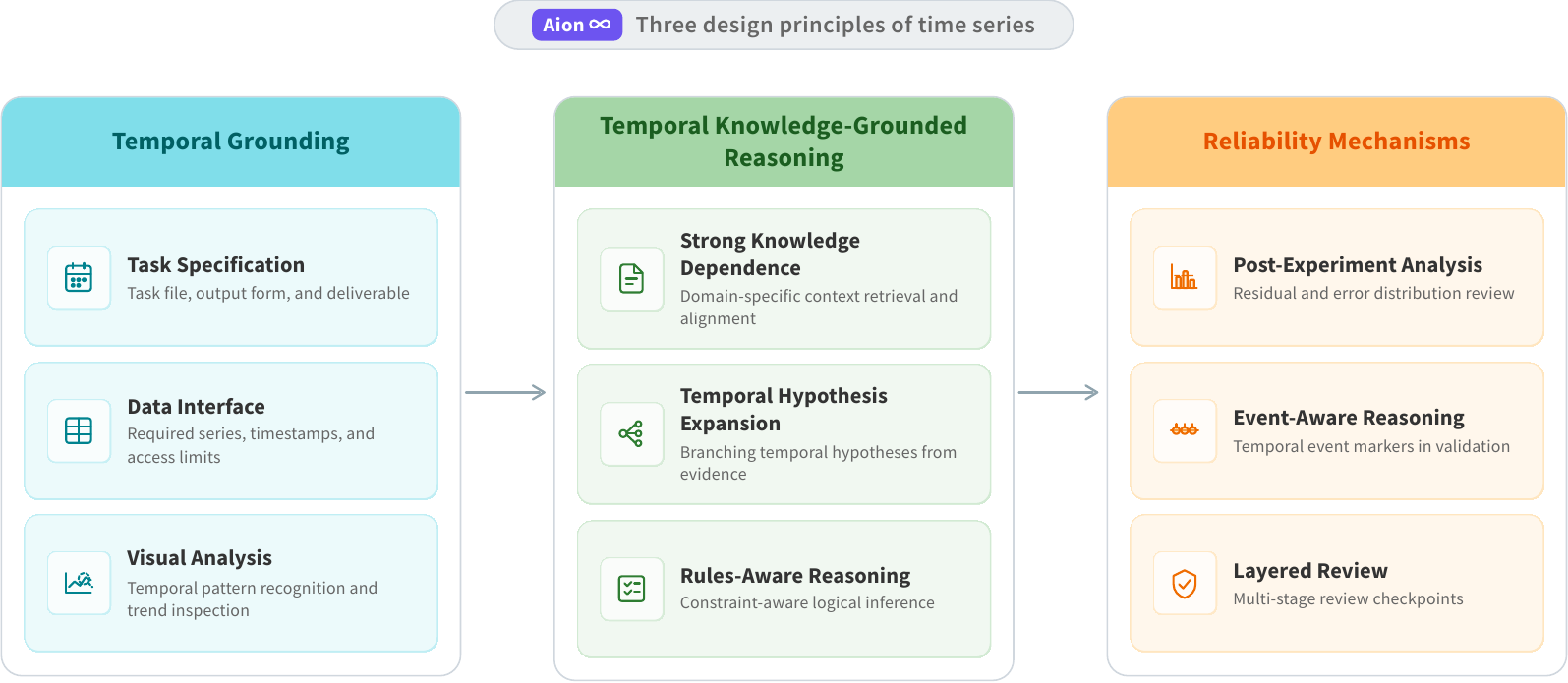}
    \caption{Overall design principles view of the time series harness. The harness is specialized through three coupled design principles: temporal grounding, temporal knowledge-grounded reasoning, and reliability mechanisms.}
    \label{fig:core-time-series-principles}
\end{figure}

We describe three design principles for time series harnesses as shown in Figure~\ref{fig:core-time-series-principles}. The following subsections detail each one.

\subsubsection{Temporal Grounding}\label{sec:temporal-grounding}
Temporal grounding concerns the parts that are closest to the time series itself. Before asking which model to use, the harness must first determine \emph{what counts as one temporal observation, which horizon is requested, and whether the visible data can support that request.} In this paper, \emph{temporal constraints} are the time-related requirements that any valid answer must satisfy, including the observation unit, allowed history and cutoff, forecast or reasoning horizon, timestamp granularity, output frequency, and leakage boundary. \emph{Temporal coverage} is the span of visible history together with the target window that the system must forecast, explain, or evaluate.

\begin{figure}[h]
    \centering
    \includegraphics[width=0.7\textwidth]{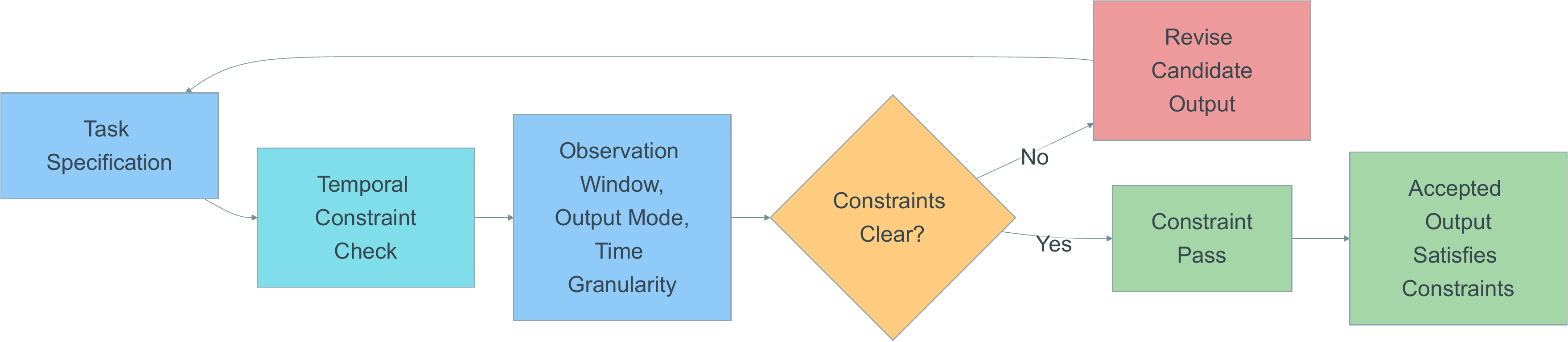}
    \caption{Task specification: a loop that determines the required observation window, output mode, and time granularity, then checks whether a candidate forecast or structured answer satisfies the temporal constraints.}
    \label{fig:tg-forecast-control}
\end{figure}

\hl{\textbf{Task Specification.}} \method first resolves the task file and validation interface defined in Section~\ref{sec:task-specification-for-next-generation-time-series-problems}. This step asks what observations the task requires, what legal output form is expected, and what final deliverable must be produced, such as a forecast table, a structured temporal answer, a report, or a platform submission. In temporal terms, these requirements become the observation window, output mode, and time granularity shown in Figure~\ref{fig:tg-forecast-control}. A candidate output is accepted only if it satisfies these constraints without crossing the leakage boundary.

\begin{painpointbox}
    The task specification stage identifies the required observations, output form, validation interface, and deliverables, then makes the temporal constraints explicit before the system is allowed to accept a forecast or structured answer.

    \ref{sec:appendix-agents} \textbf{Agents:}
    \begin{itemize}
        \item \ref{sec:appendix-agent-interpreter} \emph{Task Interpretation and Task Specification} (\hl{\emph{Interpreter}})
        \item \ref{sec:appendix-agent-temporal-governor} \emph{Time Series Review Control} (\hl{\emph{Temporal Governor}})
    \end{itemize}
    \ref{sec:appendix-skills} \textbf{Skills:}
    \begin{itemize}
        \item \ref{sec:appendix-skills-data-interface-unification} \emph{Data Interface Unification}
        \item \ref{sec:appendix-skills-forecast-output-control} \emph{Forecast Output Control}
    \end{itemize}
    \ref{sec:appendix-rules} \textbf{Rules:}
    \begin{itemize}
        \item \ref{sec:appendix-rules-time-series-constraint-checks} \emph{Time Series Constraint Checks}
    \end{itemize}
    \ref{sec:appendix-protocols} \textbf{Protocols:}
    \begin{itemize}
        \item \ref{sec:appendix-protocols-lifecycle-state-model} \emph{Lifecycle State Model}
    \end{itemize}

    Figure~\ref{fig:tg-forecast-control} summarizes this stage as a compact pass-or-revise check: the system derives the required observation window, output mode, and time granularity, then revises any candidate output that fails those constraints.
\end{painpointbox}

\hl{\textbf{Data Interface.}} Once the task requirements are clear, \method defines the data interface that will support those task requirements. Real tasks often arrive through heterogeneous files, partial tables, logs, or platform exports, so the system must decide which series, auxiliary files, timestamps, entities, and access limits are relevant to the task. Here, a stable data interface means a representation of the available series and auxiliary files with explicit timestamp granularity, variable meanings, task-series relations, and leakage-prevention constraints.

\begin{figure}[h]
    \centering
    \includegraphics[width=0.7\textwidth]{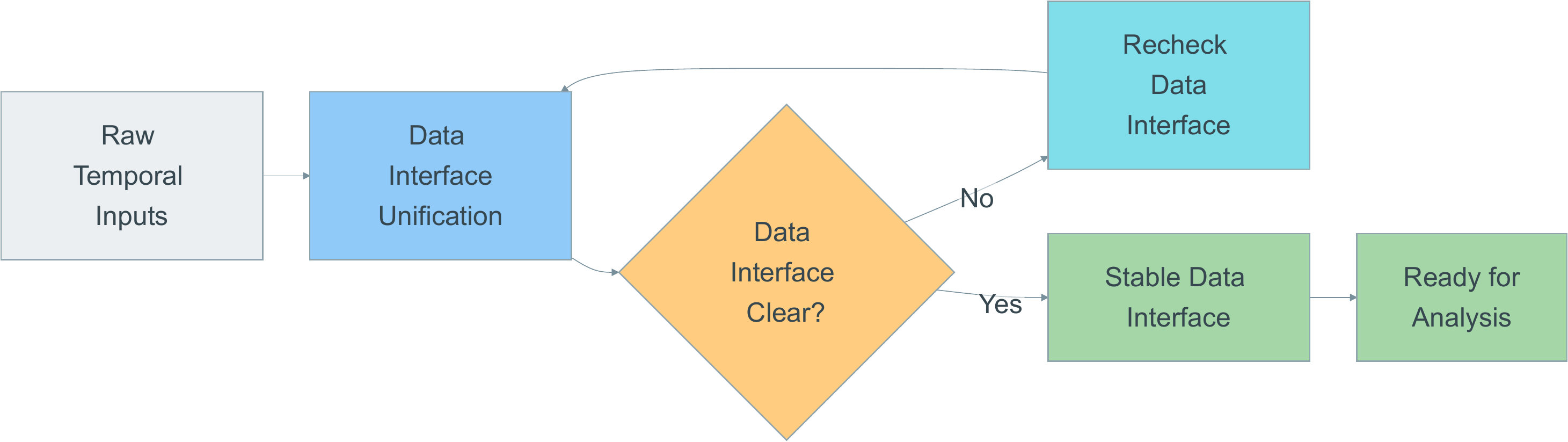}
    \caption{Data interface: from raw inputs through data interface unification to a stable representation that makes timestamp granularity, chronological semantics, variable meaning, task-series relations, and leakage-prevention constraints explicit.}
    \label{fig:tg-data-entry}
\end{figure}

\begin{painpointbox}
    The data interface definition described above is realized through the coordinated action of multiple harness components. This definition uses the task specification to decide which fields and temporal boundaries must be exposed before analysis begins.

    \ref{sec:appendix-agents} \textbf{Agents:}
    \begin{itemize}
        \item \ref{sec:appendix-agent-interpreter} \emph{Task Interpretation and Task Specification} (\hl{\emph{Interpreter}})
    \end{itemize}
    \ref{sec:appendix-skills} \textbf{Skills:}
    \begin{itemize}
        \item \ref{sec:appendix-skills-workspace-initialization-and-context-anchoring} \emph{Workspace Initialization and Context Anchoring}
        \item \ref{sec:appendix-skills-input-safety-screening} \emph{Input Safety Screening}
        \item \ref{sec:appendix-skills-document-and-attachment-intake} \emph{Document and Attachment Intake}
        \item \ref{sec:appendix-skills-data-interface-unification} \emph{Data Interface Unification}
    \end{itemize}
    \ref{sec:appendix-rules} \textbf{Rules:}
    \begin{itemize}
        \item \ref{sec:appendix-rules-execution-surface-conventions} \emph{Execution Surface Conventions}
    \end{itemize}
    \ref{sec:appendix-protocols} \textbf{Protocols:}
    \begin{itemize}
        \item \ref{sec:appendix-protocols-dispatch-interface} \emph{Dispatch Interface}
    \end{itemize}

    Figure~\ref{fig:tg-data-entry} illustrates the short path from raw temporal inputs to a stable data interface.
\end{painpointbox}

\hl{\textbf{Visual Analysis.}} \method places visual analysis before deeper reasoning. Here, visual analysis means inspecting plots or visual summaries for shape-based signals such as drift, peaks, missing values, state changes, local misalignment, or temporal boundary effects. The agent's visual capabilities (vision language models, such as \texttt{gpt-5.x}, \texttt{claude-opus-4.x}, and \texttt{qwen3.6-plus}) can capture these signals before the system decides whether the task should move toward prediction, event reasoning, anomaly analysis, or further information gathering. \textbf{In this process, visual summaries become intermediate evidence that can be checked and reused during agent-based reasoning.}

\begin{figure}[t]
    \centering
    \includegraphics[width=0.7\textwidth]{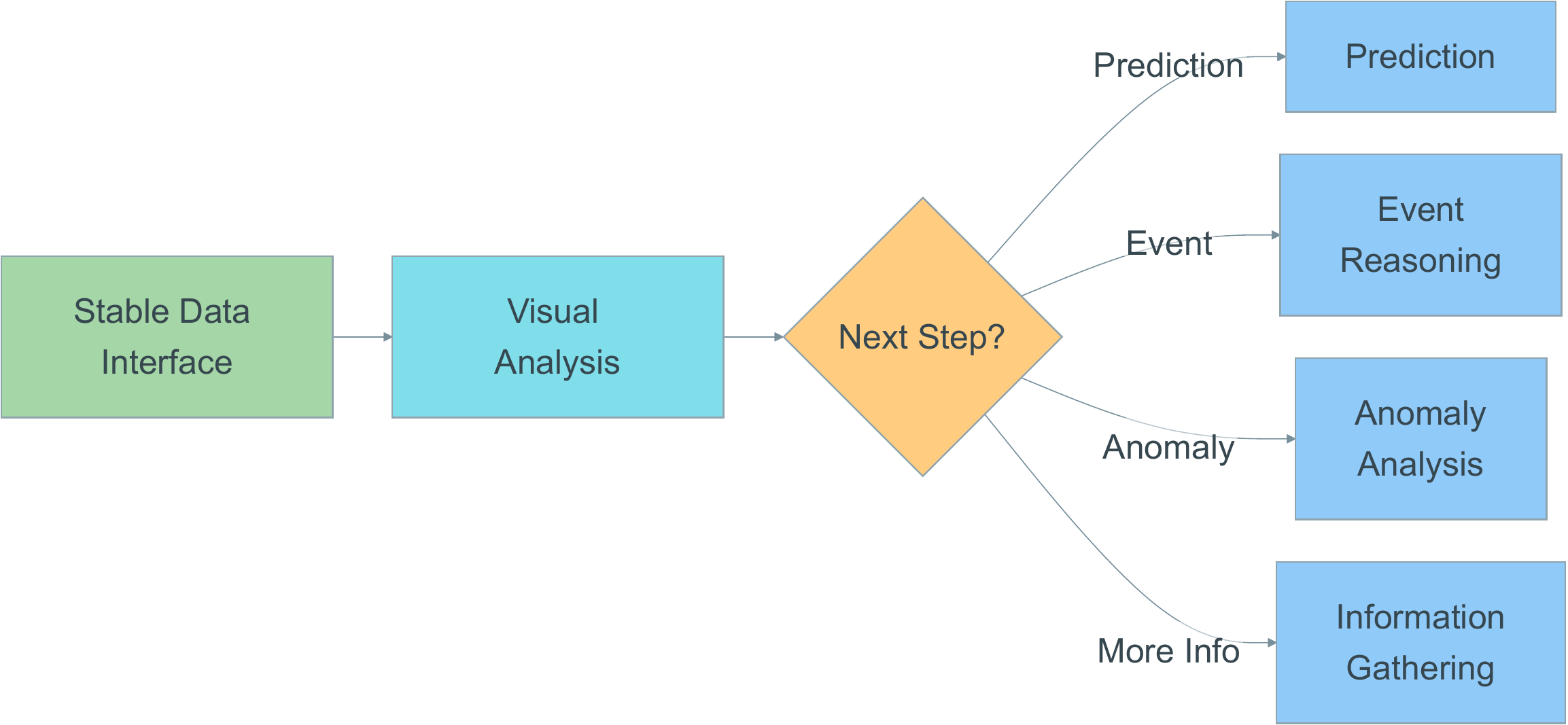}
    \caption{Visual analysis within the process branch: the selection stage captures shape-based signals (drift, peaks, missing values, state changes, local misalignment, etc.) from the stable data interface, then decides whether the task should move toward prediction, event reasoning, anomaly analysis, or further information gathering.}
    \label{fig:tg-analysis}
\end{figure}

\begin{painpointbox}
    Visual analysis is the selection stage of temporal grounding. The harness reads the visible shape first, then decides which next analysis step should run.

    \ref{sec:appendix-agents} \textbf{Agents:}
    \begin{itemize}
        \item \ref{sec:appendix-agent-evidence-collector} \emph{External Evidence Expansion} (\hl{\emph{Evidence Collector}})
    \end{itemize}
    \ref{sec:appendix-skills} \textbf{Skills:}
    \begin{itemize}
        \item \ref{sec:appendix-skills-document-and-attachment-intake} \emph{Document and Attachment Intake}
        \item \ref{sec:appendix-skills-time-series-analysis-prior} \emph{Time Series Analysis Prior}
    \end{itemize}
    \ref{sec:appendix-rules} \textbf{Rules:}
    \begin{itemize}
        \item \ref{sec:appendix-rules-execution-surface-conventions} \emph{Execution Surface Conventions}
    \end{itemize}
    \ref{sec:appendix-protocols} \textbf{Protocols:}
    \begin{itemize}
        \item \ref{sec:appendix-protocols-dispatch-interface} \emph{Dispatch Interface}
    \end{itemize}

    Figure~\ref{fig:tg-analysis} highlights the compact path from the stable data interface through visual analysis to a decision about the next analysis step.
\end{painpointbox}

\subsubsection{Temporal Knowledge-Grounded Reasoning}\label{sec:knowledge-grounded-reasoning}
Once temporal grounding is established, a time series harness should move to temporal knowledge-grounded reasoning. By this, we mean reasoning that uses task context, domain knowledge, external evidence, and platform rules to interpret the time series rather than relying on the numeric sequence alone. In real-world problems, the pure time series data is often insufficient for reliable problem-solving; the system must also interpret the context around the series. We address this through the following design consideration.

\hl{\textbf{Strong Knowledge Dependence.}} This means the series cannot be interpreted reliably from its numeric values alone. Two recurring conditions often appear together. First, the system often needs to interpret timestamp meaning, event-driven variation, seasonal recurrence, and temporal pattern shifts because these semantics are not fully specified in advance. Second, weakly validated data pipelines, meaning data loading, cleaning, or merging steps that have not yet been checked against the task rules, often require additional collection, cleaning, and interpretation before modeling. This dependence is further shaped by the specific time series domain, as well as different contextual requirements. We therefore treat time series modeling as a temporal knowledge-grounded reasoning process, which makes it a natural fit for a harness.

\begin{figure}[h]
    \centering
    \includegraphics[width=0.7\textwidth]{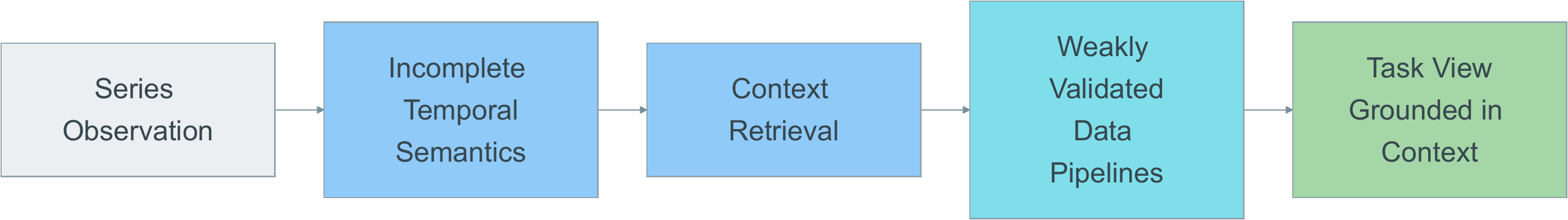}
    \caption{Strong knowledge dependence: reasoning in a time series harness can progress from series observation to incomplete temporal semantics, context retrieval, and weakly validated data pipelines, ultimately forming a task view grounded in context. This view requires the system to reason about the data environment rather than treat it as a static input interface.}
    \label{fig:kg-strong-knowledge}
\end{figure}

\begin{painpointbox}
    This requires the harness to reason about the time-indexed data environment rather than treat it as a fixed input interface.

    \ref{sec:appendix-agents} \textbf{Agents:}
    \begin{itemize}
        \item \ref{sec:appendix-agent-interpreter} \emph{Task Interpretation and Task Specification} (\hl{\emph{Interpreter}})
        \item \ref{sec:appendix-agent-evidence-collector} \emph{External Evidence Expansion} (\hl{\emph{Evidence Collector}})
    \end{itemize}
    \ref{sec:appendix-skills} \textbf{Skills:}
    \begin{itemize}
        \item \ref{sec:appendix-skills-document-and-attachment-intake} \emph{Document and Attachment Intake}
        \item \ref{sec:appendix-skills-data-interface-unification} \emph{Data Interface Unification}
        \item \ref{sec:appendix-skills-time-series-analysis-prior} \emph{Time Series Analysis Prior}
    \end{itemize}
    \ref{sec:appendix-protocols} \textbf{Protocols:}
    \begin{itemize}
        \item \ref{sec:appendix-protocols-dispatch-interface} \emph{Dispatch Interface}
    \end{itemize}

    Figure~\ref{fig:kg-strong-knowledge} summarizes this stage as a path from observed series to a task view grounded in context, after context retrieval and data-pipeline checking.
\end{painpointbox}

\hl{\textbf{Temporal Hypothesis Expansion.}} A temporal hypothesis is a candidate explanation of an observed pattern in terms of temporal dynamics, such as an event effect, a transient shock, a seasonal carryover, or a regime shift.
Because the same observations may admit several such explanations, the harness should not commit to one early. It therefore first expands candidate temporal hypotheses over temporal scale, event alignment, lag structure, seasonal recurrence, and regime segmentation, then tests them against the available evidence before narrowing to one.

\begin{figure}[h]
    \centering
    \includegraphics[width=0.7\textwidth]{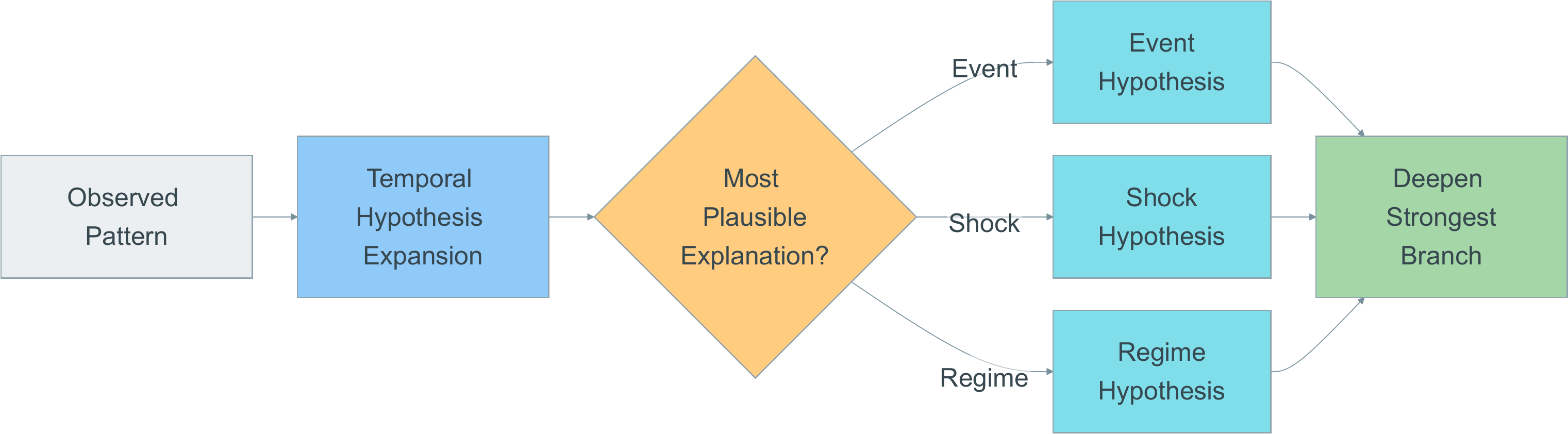}
    \caption{Temporal hypothesis expansion within the process branch: observed patterns are expanded into event, shock, and regime hypotheses before one candidate hypothesis is selected for deeper analysis.}
    \label{fig:kg-hypothesis-expansion}
\end{figure}

\begin{painpointbox}
    Here, the harness treats candidate hypotheses as branches that must be widened before one candidate hypothesis is allowed to dominate, especially when multiple time-based mechanisms remain plausible.

    \ref{sec:appendix-agents} \textbf{Agents:}
    \begin{itemize}
        \item \ref{sec:appendix-agent-orchestrator} \emph{Global Orchestration and Task Integration} (\hl{\emph{Orchestrator}})
        \item \ref{sec:appendix-agent-evidence-collector} \emph{External Evidence Expansion} (\hl{\emph{Evidence Collector}})
    \end{itemize}
    \ref{sec:appendix-skills} \textbf{Skills:}
    \begin{itemize}
        \item \ref{sec:search-space-expansion} \emph{Search Space Expansion}
        \item \ref{sec:appendix-skills-long-horizon} \emph{Long Horizon Reasoning and Dependency Tightening}
    \end{itemize}
    \ref{sec:appendix-rules} \textbf{Rules:}
    \begin{itemize}
        \item \ref{sec:search-and-source-expansion-policy} \emph{Search and Source Expansion Policy}
    \end{itemize}

    Figure~\ref{fig:kg-hypothesis-expansion} shows this stage as expanding event, shock, and regime hypotheses before deepening the strongest branch.
\end{painpointbox}

\hl{\textbf{Rules-Aware Reasoning.}} For tasks with strict rules, such as public benchmarks or competition hosts, the system reasons explicitly about leaderboard structure, submission constraints, and high-scoring solution patterns. In real-world time series tasks, this often includes forecast horizons, submission cutoffs, update cadence, and other platform-level temporal constraints that shape the solution.

\begin{figure}[h]
    \centering
    \includegraphics[width=0.7\textwidth]{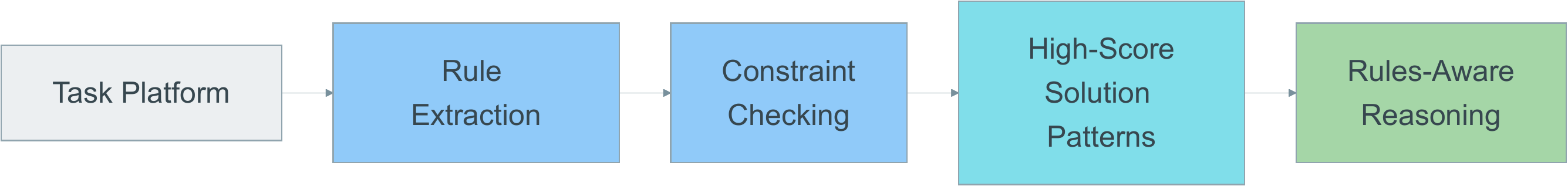}
    \caption{Rules-aware reasoning within the process branch: from the task platform through rule extraction and constraint checking to high-score solution patterns, shaping a rules-aware reasoning path where external heuristics are treated as evidence.}
    \label{fig:kg-rules-aware}
\end{figure}

\begin{painpointbox}
    When the task lives inside an external ruleset, the harness treats that ruleset as evidence, especially when the rules constrain temporal coverage or forecast horizon.

    \ref{sec:appendix-agents} \textbf{Agents:}
    \begin{itemize}
        \item \ref{sec:appendix-agent-evidence-collector} \emph{External Evidence Expansion} (\hl{\emph{Evidence Collector}})
        \item \ref{sec:appendix-agent-constructor} \emph{Solution Construction and Experimental Realization} (\hl{\emph{Constructor}})
    \end{itemize}
    \ref{sec:appendix-skills} \textbf{Skills:}
    \begin{itemize}
        \item \ref{sec:appendix-skills-tool-prior} \emph{Tool Prior and Analysis Toolkit}
        \item \ref{sec:public-artifact-evidence-retrieval} \emph{Public Artifact Evidence Retrieval}
    \end{itemize}
    \ref{sec:appendix-rules} \textbf{Rules:}
    \begin{itemize}
        \item \ref{sec:search-and-source-expansion-policy} \emph{Search and Source Expansion Policy}
        \item \ref{sec:appendix-rules-execution-surface-conventions} \emph{Execution Surface Conventions}
    \end{itemize}
    \ref{sec:appendix-protocols} \textbf{Protocols:}
    \begin{itemize}
        \item \ref{sec:appendix-protocols-runtime-event-logging} \emph{Runtime Event Logging}
    \end{itemize}

    Figure~\ref{fig:kg-rules-aware} summarizes this stage as extracting platform rules, reading public heuristics, and shaping a rules-aware reasoning path.
\end{painpointbox}

\subsubsection{Reliability Mechanisms}\label{sec:reliability-mechanisms}
The final principle concerns system reliability. Even temporally grounded, temporal knowledge-grounded reasoning can still fail if they accept weak evidence, overcommit to branches early, or stop before validations. Reliable execution, therefore, requires explicit mechanisms for evidence control, branch management, and claim validation. We emphasize three core reliability mechanisms here.

\hl{\textbf{Post-Experiment Analysis.}} After a run finishes, the system analyzes why the result succeeded or failed through error decomposition, failure case studies, feature attribution, and residual checks.

\begin{figure}[h]
    \centering
    \includegraphics[width=0.7\textwidth]{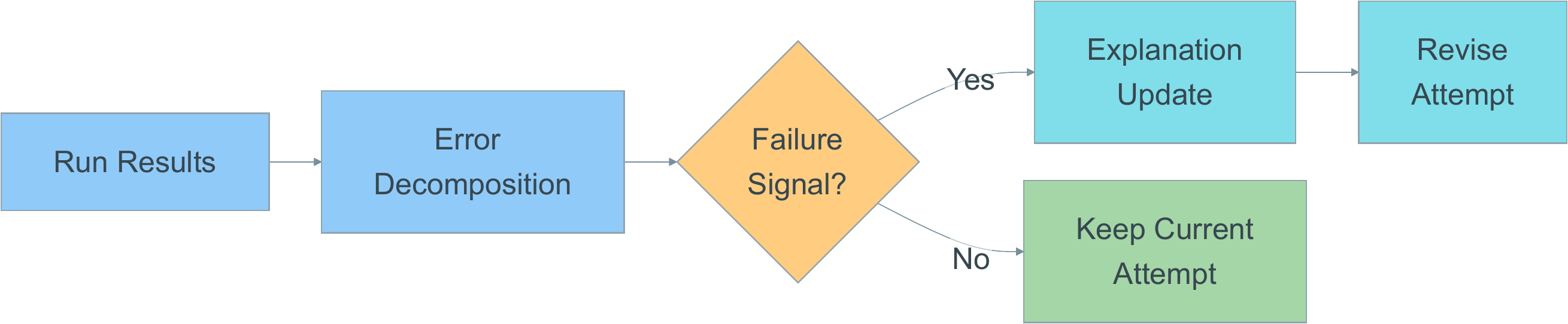}
    \caption{Post-experiment analysis: run results are inspected through error decomposition and failure signal detection. Detected failures update the explanation of the result and revise the current solution attempt, while runs without such signals keep the current attempt.}
    \label{fig:lr-post-experiment}
\end{figure}

\begin{painpointbox}
    After a run finishes, this mechanism checks whether the result still contains unresolved errors. If the errors point to a weak explanation of the result, the system updates that explanation and revises the current solution attempt before the next round.

    \ref{sec:appendix-skills} \textbf{Skills:}
    \begin{itemize}
        \item \ref{sec:appendix-skills-critical-rollback-and-gap-review} \emph{Critical Rollback and Gap Review}
    \end{itemize}
    \ref{sec:appendix-evaluation} \textbf{Evaluation:}
    \begin{itemize}
        \item \ref{sec:appendix-evaluation-regression-comparison-matrix} \emph{Regression Comparison Matrix}
    \end{itemize}

    Figure~\ref{fig:lr-post-experiment} shows this stage as a short loop from result analysis to error decomposition, failure signal detection, explanation update, and solution revision.
\end{painpointbox}

\hl{\textbf{Event-Aware Reasoning.}} Event-aware reasoning checks whether an observed change should be explained by an event, a local shock, or neither. The harness frames candidate events, identifies temporal boundaries and effect durations, and analyzes whether events alter levels, volatility, relational structures, or regimes.

\begin{figure}[h]
    \centering
    \includegraphics[width=0.7\textwidth]{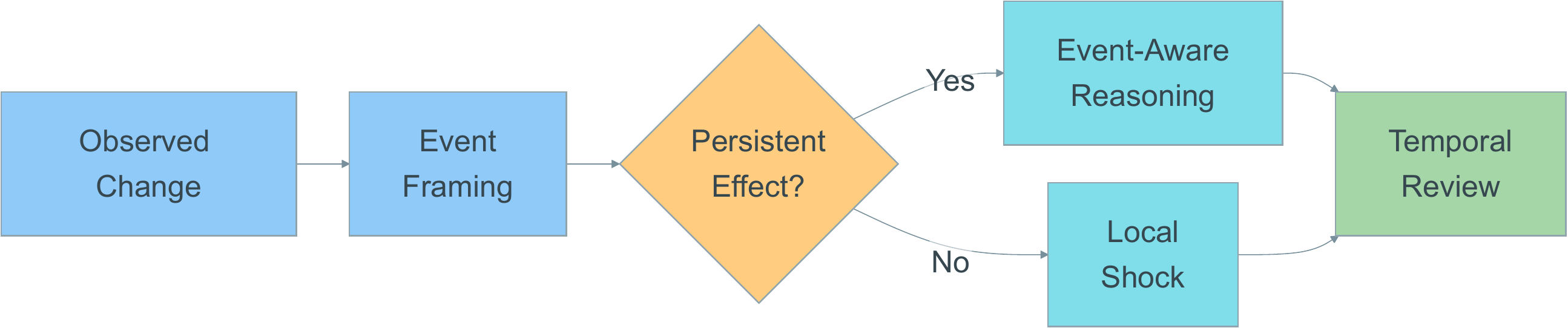}
    \caption{Event-aware reasoning within the process branch: from observed change through event framing and persistent effect detection to event-aware reasoning or local shock, converging at temporal review, where temporal boundaries and effect durations are evaluated to determine whether events alter levels, volatility, relational structures, or regimes.}
    \label{fig:lr-event-aware}
\end{figure}

\begin{painpointbox}
    Event-aware reasoning remains reliable only when event framing, temporal constraints, and runtime traces stay aligned. Event framing means linking an observed change to a possible event and checking whether the effect persists beyond a local shock.

    \ref{sec:appendix-agents} \textbf{Agents:}
    \begin{itemize}
        \item \ref{sec:appendix-agent-temporal-governor} \emph{Time Series Review Control} (\hl{\emph{Temporal Governor}})
    \end{itemize}
    \ref{sec:appendix-skills} \textbf{Skills:}
    \begin{itemize}
        \item \ref{sec:appendix-skills-long-horizon} \emph{Long Horizon Reasoning and Dependency Tightening}
    \end{itemize}
    \ref{sec:appendix-protocols} \textbf{Protocols:}
    \begin{itemize}
        \item \ref{sec:appendix-protocols-runtime-event-logging} \emph{Runtime Event Logging}
    \end{itemize}

    Figure~\ref{fig:lr-event-aware} highlights the compact event-aware path from observed change through event framing and persistent effect checking to temporal review.
\end{painpointbox}

\hl{\textbf{Layered Review.}} The system inspects critical steps before and after execution, allowing the harness to reject weak candidate solutions early, send insufficient candidate solutions into a recheck loop, and prevent superficial progress from being mistaken for completion. In time series tasks, this is where forecast horizon, leakage boundaries, and temporal alignment are repeatedly rechecked.

\begin{figure}[h]
    \centering
    \includegraphics[width=0.7\textwidth]{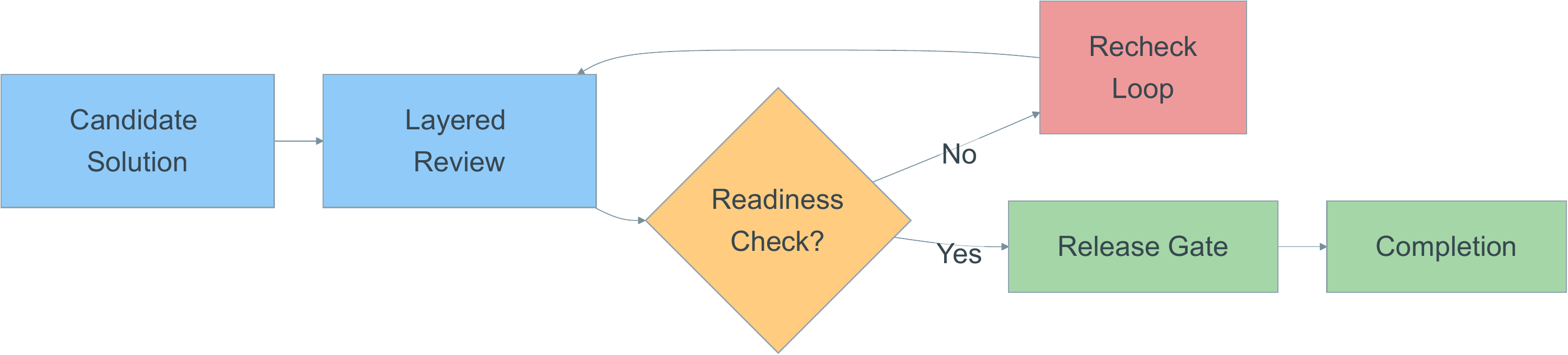}
    \caption{Layered review within the process branch: from candidate solution through layered review to readiness check, with a recheck loop on rejection and release gate on acceptance leading to completion.}
    \label{fig:lr-layered-review}
\end{figure}

\begin{painpointbox}
    In this design, a layered review serves as a review mechanism throughout execution and completion. A readiness check asks whether the candidate solution has enough evidence to continue toward completion; a release gate is the final acceptance check before the system stops.

    \ref{sec:appendix-agents} \textbf{Agents:}
    \begin{itemize}
        \item \ref{sec:appendix-agent-final-reviewer} \emph{Minimal Context Final Recheck} (\hl{\emph{Final Reviewer}})
    \end{itemize}
    \ref{sec:appendix-evaluation} \textbf{Evaluation:}
    \begin{itemize}
        \item \ref{sec:appendix-evaluation-release-readiness-gates} \emph{Release Readiness Gates}
    \end{itemize}
    \ref{sec:appendix-protocols} \textbf{Protocols:}
    \begin{itemize}
        \item \ref{sec:appendix-protocols-review-response-and-recheck-interface} \emph{Review Response and Recheck Interface}
    \end{itemize}

    Figure~\ref{fig:lr-layered-review} summarizes layered review as a review point before and after execution that either allows release or sends the candidate solution into a recheck loop.
\end{painpointbox}

These three design principles form the time series backbone of \method. The next section demonstrates these mechanisms in a concrete Kaggle competition setting.

\section{Kaggle Practical Case on Store Sales Competition}\label{sec:kaggle-practical-case-store-sales-competition}

\subsection{Competition Background}\label{sec:competition-background}

We use the Kaggle competition \textit{Store Sales - Time Series Forecasting}\footnote{\url{https://www.kaggle.com/competitions/store-sales-time-series-forecasting}} as a case study in this technical report. The competition asks participants to forecast future retail sales from historical store-level records and related business context, then submit predictions in a standard Kaggle format. Although still a forecasting task, this competition already exhibits the structural elements of our formulation: an explicit horizon, auxiliary context, submission rules, and leaderboard-based evaluation. Under Definition~\ref{def:ngts-task} and Eq.~\eqref{eq:ngts-task-triple}, the competition can be viewed as a concrete task object $\mathcal{T} = (\phi, \mathcal{W}, \mathcal{V})$.

\begin{itemize}
    \item \textbf{Task file $\phi$}: The comprehensive specification of the competition.
          \begin{itemize}
              \item $\mathbb{T}$ establishes the access scope, including the admissible time span and the prediction horizon.
                    \begin{painpointbox}
                        The official task asks for forecasts from 2017-08-16 to 2017-08-31, i.e.\ a fixed 16-day horizon.
                    \end{painpointbox}
              \item $\Sigma$ captures the prior knowledge relevant to the task.
                    \begin{painpointbox}
                        Relevant prior knowledge includes grocery retail seasonality, promotional effects, holidays, and store-level business contexts, extending beyond simple univariate extrapolation.
                    \end{painpointbox}
              \item $\mathcal{O}$ defines the output form.
                    \begin{painpointbox}
                        Participants must upload predictions in the standardized Kaggle submission format rather than return free-form reports.
                    \end{painpointbox}
              \item $\mathcal{C}$ collects hard requirements such as submission and leakage constraints.
                    \begin{painpointbox}
                        A forecast is considered meaningful only if it satisfies platform-level submission validity and adheres to the designated train-test split.
                    \end{painpointbox}
              \item $A$ collects the performance evaluation metrics used for ranking.
                    \begin{painpointbox}
                        The official score uses RMSLE:
                        \begin{equation}\label{eq:kaggle-rmsle}
                            \mathrm{RMSLE}(y_i, \hat{y}_i) = \sqrt{\frac{1}{n}\sum_{i=1}^{n}{\left(\log(\hat{y}_i+1)-\log(y_i+1)\right)}^2},
                        \end{equation}
                        Leaderboard ranking is determined by this scalar metric.
                    \end{painpointbox}
          \end{itemize}
    \item \textbf{Workspace $\mathcal{W}$}: The set of resources that a participant is permitted to utilize throughout the competition.
          \begin{itemize}
              \item The workspace comprises the released data files rather than a single, processed table.
                    \begin{painpointbox}
                        Kaggle provides multiple files, including \texttt{train.csv}, \texttt{test.csv}, and several auxiliary context tables. Consequently, the solver operates over a set of interrelated artifacts rather than processing a single, isolated data series.
                    \end{painpointbox}
              \item The workspace also includes metadata-bearing column descriptions and auxiliary artifacts used to interpret these files.
                    \begin{painpointbox}
                        Column identifiers, auxiliary tables, and released metadata help determine how the forecasting target is reconstructed from the available artifacts.
                    \end{painpointbox}
              \item The workspace may reveal only a subset of the underlying business state.
                    \begin{painpointbox}
                        Participants observe only the released tables and metadata rather than the complete latent operational state of the retailer. Consequently, the competition provides only a partial representation of the actual system.
                    \end{painpointbox}
              \item The workspace includes executable tools within the environment for training, validation, and submission generation.
                    \begin{painpointbox}
                        Effective participation depends on the implementation of executable code for feature construction, model fitting, local validation, and the packaging of submissions, rather than relying exclusively on a static prediction table.
                    \end{painpointbox}
          \end{itemize}
    \item \textbf{Validation interface $\mathcal{V}$}: The evaluation pipeline established for the competition.
          \begin{itemize}
              \item A candidate output must first satisfy the validity checks collected in $g$.
                    \begin{painpointbox}
                        An uploaded file must fulfill the requirements of a valid submission before Kaggle performs the scoring procedure.
                    \end{painpointbox}
              \item The submission is subsequently evaluated based on the scoring metrics defined in $s$.
                    \begin{painpointbox}
                        The valid submission is evaluated on held-out targets by the official competition metric.
                    \end{painpointbox}
          \end{itemize}
\end{itemize}

\subsection{Experimental Setup}\label{sec:experimental-setup}

We compare two execution runs on the same reconstructed Store Sales task. To avoid Kaggle submission limits while keeping a scored evaluation interface, we rebuild the competition as a local forecasting task. The local task preserves the official 16-day horizon length but shifts the hidden window earlier, from 2017-08-16--2017-08-31 to 2017-07-31--2017-08-15, so that hidden labels are available for local scoring. Both runs use the same local server, task file, and base model, \texttt{qwen3.6-plus}. The difference is in the trace. The \directbuild execution uses the OpenCode agent\footnote{OpenCode is an open-source AI coding agent (\url{https://opencode.ai/}).} in direct-build mode and quickly moves to a feature-engineered \texttt{LightGBM} submission path. In contrast, the \method execution begins with context recovery, exploratory analysis, and baseline comparison, including a recent-week median baseline.

\begin{painpointbox}
    \textbf{Reconstructed task context}
    \begin{itemize}
        \item \textbf{Local slice:} The local slice maintains stores 1 through 5 and retains all product families. It exposes \texttt{train.csv} exclusively until 2017-07-30, whereas the period from 2017-07-31 to 2017-08-15 is designated as the hidden forecast window.
        \item \textbf{Prediction object:} The public \texttt{test.csv} is constructed from the held-out rows following the removal of the target column. The resulting task preserves the original granularity while incorporating a fixed 16-day horizon and 2640 evaluation rows.
        \item \textbf{Feature boundary:} The \texttt{transactions.csv} file is truncated at the boundary of the public training set, whereas \texttt{oil.csv} and \texttt{holidays\_events.csv} remain accessible throughout the hidden window.
        \item \textbf{Server loop:} A lightweight local HTTP service emulates the core Kaggle workflow: file discovery, file download, raw CSV submission, scoring against hidden ground truth, submission history, and leaderboard feedback.
        \item \textbf{Task file:} The associated task file defines the target, temporal coverage, required columns, temporal constraints, RMSLE metric, and concrete HTTP endpoints, so each execution run starts from the same executable task specification.
    \end{itemize}
\end{painpointbox}

\subsection{Trace Overview}\label{sec:trace-overview}

\begin{figure}[h]
    \centering
    \includegraphics[width=0.95\textwidth]{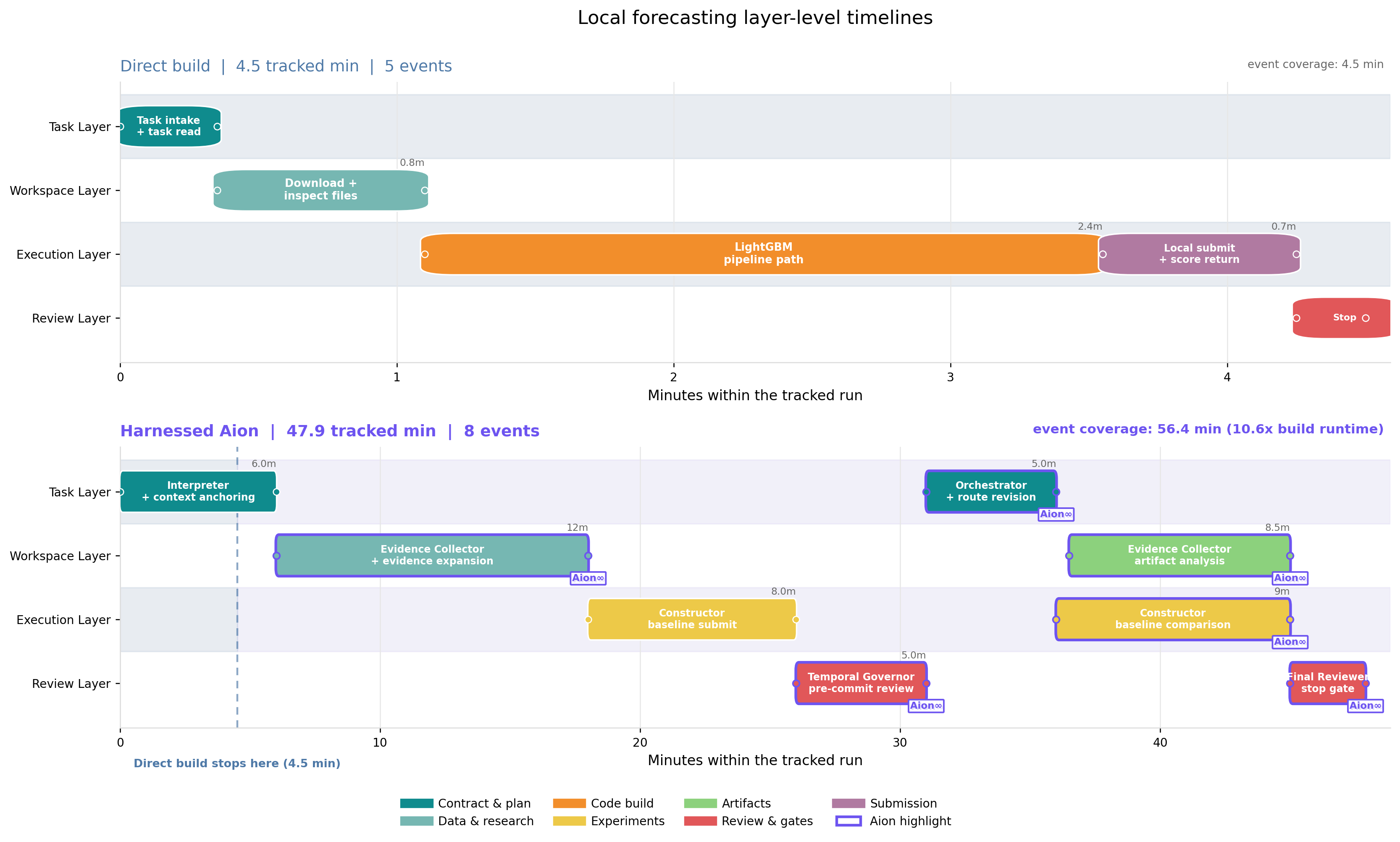}
    \caption{Phase-level milestones for the \directbuild run and the harnessed \method run. The direct run stays short and linear, moving from task intake to file inspection, a single LightGBM submission path, and an external stop. In contrast, the harnessed run adds explicit context anchoring, evidence expansion, baseline submission and comparison, solution revision, artifact analysis, and completion after review.}
    \label{fig:kaggle-process-timeline}
\end{figure}

Figure~\ref{fig:kaggle-process-timeline} illustrates the same contrast at the process level. The \directbuild execution follows a concise linear pipeline: it accesses the task, examines files, adopts a single \texttt{LightGBM} strategy, retrieves a scored submission, and terminates externally. In contrast, the \method execution revisits context, expands evidence, compares strategies, generates intermediate artifacts, and reviews the result before termination. This extra work is spent on self-correction, strategy verification, and completion checks.

\begin{figure}[h]
    \centering
    \includegraphics[width=0.95\textwidth]{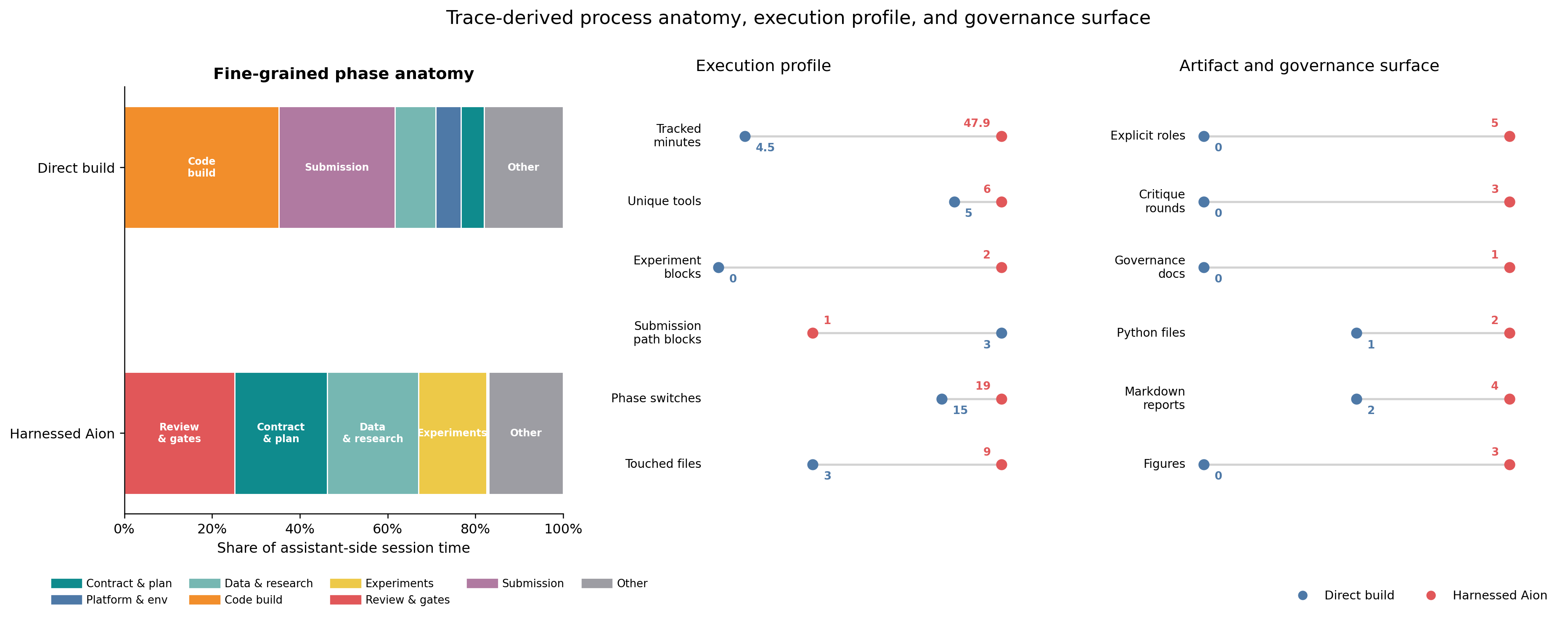}
    \caption{Trace-derived process statistics. The left panel breaks down time allocation across execution phases, the middle panel compares operational effort, and the right panel compares artifact output and review activity. Together, these panels show where the harness invests additional computation relative to the shorter \directbuild run, complementing the RMSLE comparison in Table~\ref{tab:kaggle-case-results}.
    }
    \label{fig:kaggle-process-stats}
\end{figure}

Figure~\ref{fig:kaggle-process-stats} summarizes the same contrast with aggregate process statistics. The goal is to show where the extra computation goes. Compared with the shorter \directbuild trace, \method spends more tracked time and operations on analysis, review, and artifact generation. These stages are where strategy selection and completion checks appear in the trace. Table~\ref{tab:kaggle-case-results} reports the final scored submission together with trace-level statistics for each run. We interpret these results in the following discussion.

\begin{table}[h]
    \centering
    \caption{Recorded outcome summary for the local Kaggle case. Lower RMSLE is better.}
    \label{tab:kaggle-case-results}
    \renewcommand{\arraystretch}{1.14}
    \begin{tabular}{ccc}
        \toprule
        Measure                & \directbuild & Harnessed \method \\
        \midrule
        Final submission RMSLE & 1.6811       & 0.4352            \\
        \midrule
        Tracked runtime (min)  & 4.5          & 47.9              \\
        Unique tools           & 5            & 6                 \\
        Touched files          & 3            & 9                 \\
        Review blocks          & 0            & 3                 \\
        Critique rounds        & 0            & 3                 \\
        Figures produced       & 0            & 3                 \\
        \bottomrule
    \end{tabular}
\end{table}

\subsection{\method Harness Effects in the Case Study}\label{sec:method-harness-effects-in-the-case-study}

This case study is not only about \method achieving a lower RMSLE than the \directbuild execution. Its main purpose is to show how the harness changes the solution organization. The \directbuild execution commits rapidly to a single modeling trajectory, whereas the harnessed run \textbf{keeps task interpretation, temporal grounding, evidence use, and review active for longer}. This makes the case study evidence about workflow structure rather than a comparison of unrelated predictive models.

\subsubsection{Execution Control Between Short Linear Build and Controlled Iteration}\label{sec:scheduling-logic-short-linear-build-vs-controlled-iteration}

Table~\ref{tab:kaggle-case-results} gives the clearest contrast between the two approaches. The \directbuild execution finishes in 4.5 minutes, modifies 3 files, and records no critique rounds or review blocks. The harnessed execution runs for 47.9 minutes, modifies 9 files, and goes through 3 critique rounds with 3 review interruptions. Figure~\ref{fig:kaggle-process-timeline} shows the same process contrast. The \directbuild moves from task acquisition to file download, follows a single \texttt{LightGBM} path, performs one submission, and stops. The session trace supports this process: the system reads the task, downloads the files, constructs a single forecasting script, submits once, achieves $\text{RMSLE} = 1.6811$, and stops.

This comparison also makes the cost clear: \method takes longer. If two runs produced outputs of the same quality, the shorter run would be preferable. Here, however, the extra time is not idle waiting. It is spent on evidence expansion, strategy checking, and review, which also increases the number of generated artifacts from 3 to 9 files. \textbf{The main point is that the framework directs additional computational resources toward self-correction and review mechanisms.}

\method spends the extra time differently. Figure~\ref{fig:kaggle-process-timeline} shows context anchoring, evidence expansion, baseline submission, strategy revision, post-experiment analysis, and final review before completion. Figure~\ref{fig:kaggle-process-stats} further shows that the added effort mainly goes to analysis, experimentation, and review, not just code generation. This means that \directbuild follows a short sequence of operations, while \method keeps interpretation, comparison, and completion checks active throughout the run.

\subsubsection{Temporal Grounding Under Explicit Temporal Constraints}\label{sec:temporal-grounding-under-explicit-temporal-constraints}

The second distinction is temporal grounding. In the reconstructed task, the forecasting window covers 2017-07-31 to 2017-08-15, while the observable training history ends on 2017-07-30. These boundaries define which evidence is allowed. The \method execution keeps these dates explicit from the start. Its initial exploratory analysis checks the training range, the testing range, and the 2640-row target before strategy expansion. In other words, the system follows the temporal constraints described in Section~\ref{sec:temporal-grounding}. The process first stabilizes the temporal window, the observable history, and the output surface.

The clearest example appears in the execution trace around a baseline that uses the corresponding weekday from the previous week. The trace records that a one-week lag is directly available only for the first seven days of the forecast period. For August 7 to August 15, the required lag values would come from July 31 to August 6, which is inside the hidden forecast window. The framework, therefore, marks that strategy as invalid and adds fallback logic before continuing. The \directbuild execution also processes dates and builds temporal features such as day, month, week, and ordinal date, but it does not turn the forecast boundary into an explicit constraint check. The core distinction is that \method does not only use temporal features; it uses the forecast horizon and history boundary to govern strategy selection and execution.

\subsubsection{Knowledge-Grounded Reasoning Before Strategy Commitment}\label{sec:knowledge-grounded-reasoning-before-strategy-commitment}

The third distinction concerns knowledge-grounded reasoning. The \directbuild execution mainly reasons within the released files and quickly moves toward a familiar modeling strategy. \method treats the task as an open evidence problem. Its session rules call for \emph{External Evidence Expansion}, \emph{Public Artifact Evidence Retrieval}, and \emph{Search Space Expansion} when a task comes from a benchmark host. In the trace, this appears as a separate evidence expansion stage before strategy commitment, as shown in Figure~\ref{fig:kaggle-process-timeline}, together with more task specification and data analysis in Figure~\ref{fig:kaggle-process-stats}. This matches Section~\ref{sec:knowledge-grounded-reasoning}: \method \textbf{does not assume the observable table is the entire problem}.

The expanded evidence base changes model selection. \method starts with basic baselines, checks that baseline evidence is informative, and keeps two alternatives for comparison. One uses the corresponding weekday from the previous week; the other uses simple exponential smoothing. The framework does not treat sophisticated models as the default starting point. The best result is the \textbf{recent-week median baseline with $\text{RMSLE} = 0.4352$}, while \directbuild commits early to a \texttt{LightGBM} path and scores 1.6811, as shown in Table~\ref{tab:kaggle-case-results}. External evidence, platform constraints, and baseline priors are therefore considered before choosing a strategy, allowing the framework to select simpler strategies when they better fit the task.

\subsubsection{Reliability Mechanisms Before Stop}\label{sec:reliability-mechanisms-before-stop}

The fourth distinction concerns reliability. The \directbuild execution treats the first scored submission as the stopping point. Table~\ref{tab:kaggle-case-results} shows no review blocks or critique rounds. Figure~\ref{fig:kaggle-process-timeline} shows the same pattern: one \texttt{LightGBM} path, one submission, and then immediate stop. Although the run retrieves a score, it does not show post-experiment analysis, blocker identification, or independent verification before stopping.

\method stops differently. Table~\ref{tab:kaggle-case-results} records three review blocks and three critique rounds. Figure~\ref{fig:kaggle-process-timeline} shows these events through temporal review, post-experiment analysis after the baseline comparison, and final review before completion. The session trace also shows the control logic in text. After the first baseline submission, plan alignment, critic review, and blocker tracking remain active. Even after validation artifacts are generated, the system still requires approval from the \hl{\emph{Temporal Governor}} and the \hl{\emph{Final Reviewer}} before final commitment. This supports the reliability view in Section~\ref{sec:reliability-mechanisms}: \method \textbf{does not treat a performance score as enough}; it requires post-experiment analysis and layered review before stopping.

\section{Conclusion}\label{sec:conclusion}

Time series research needs tasks that incorporate constraints, evidence, and validation interfaces, not only datasets and metrics.
This work formalizes next-generation time series tasks and introduces \method, a specialized harness built from six coordinated component groups.
The harness unifies temporal grounding, temporal knowledge-grounded reasoning, and reliability mechanisms into one governed execution process.
A Kaggle Store Sales case study confirms that this design produces richer process traces, more artifacts, and more explicit reviews than an unharnessed baseline. Future benchmarks should move beyond isolated dataset-metric pairs toward constraint-aware task formulations, and future agent systems for time series should adopt explicit harness structures to keep temporal grounding, evidence checking, and review active throughout execution.

\bibliographystyle{assets/plainnat}
\bibliography{main}

\clearpage
\newpage
\appendix

\input{appendix}

\end{document}

%% file: appendix.tex
\startcontents[sections]
\printcontents[sections]{l}{1}{
    \setcounter{tocdepth}{3}
    \section*{\color{ztxtechpurple} Table of Contents}
}

\section{Why \method}\label{sec:appendix-why-aion}

The name \method is not just a label. It names the initial motivation behind the harness. Time series agents should not be treated as one-shot answer generators that produce a forecast, a plot, or a report and then immediately finalize the task. They should operate across duration, under constraints that may need to be revisited, with memory of earlier assumptions and explicit opportunities for review. \method refers to \texttt{Aion}, the Hellenistic figure of unbounded, cyclical, and eternal time, often associated with the circle of the universe and the zodiac.\footnote{See the overview of Aion as a Hellenistic deity associated with perpetual, cyclic time and the zodiac, available at \url{https://en.wikipedia.org/wiki/Aion_(deity)}.} This reference matches the system goal of a governed loop that can keep checking, revising, and continuing across time.

\section{Components of \method}\label{sec:appendix-components}
\subsection{Mindmap of \method Design}\label{sec:appendix-mindmap}

\begin{figure}[h]
    \centering
    \includegraphics[width=0.8\textwidth]{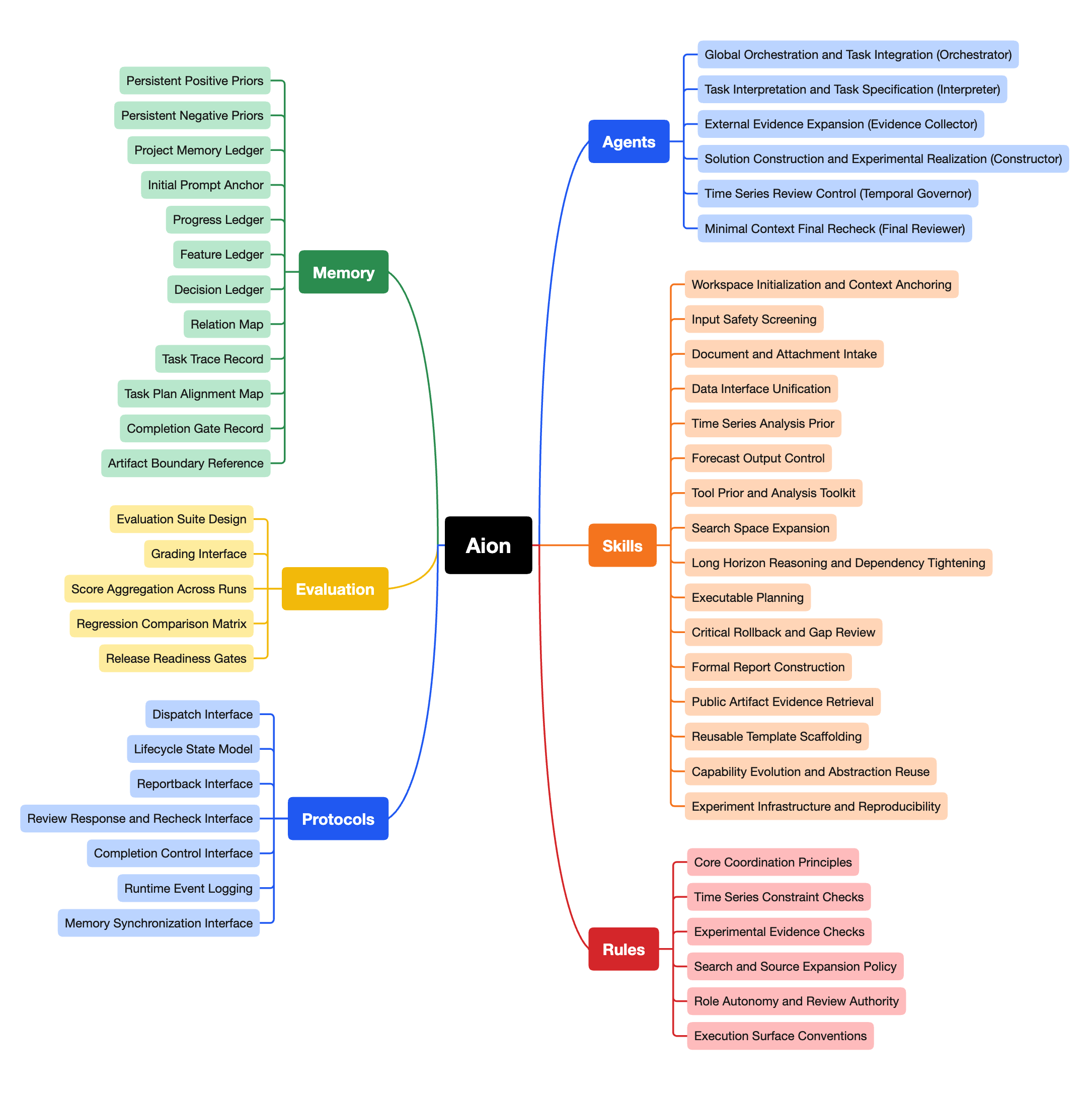}
    \caption{Mindmap of the conceptual structure behind \method.}
    \label{fig:methodmindmap}
\end{figure}

The main text introduces \method at the level of task formulation, specialization, and reliability logic. The appendix now moves one level downward and explains how that abstraction is organized into paper-level components. Rather than listing every repository file one by one, we regroup the implementation into six conceptual blocks, namely \emph{agents}, \emph{skills}, \emph{rules}, \emph{memory}, \emph{evaluation}, and \emph{protocols}. These blocks correspond to the implementation repository structure, but some appendix entries summarize multiple concrete skill or memory files under one paper-level function. This translation matters because the underlying system is granular, while the research question concerns how these pieces jointly realize a controllable time series harness. Figure~\ref{fig:methodmindmap} is a compact structural guide for the remainder of this appendix, and the following subsections unpack each block in turn.

\subsection{Agents}\label{sec:appendix-agents}

In \method, agents are not introduced as a collection of interchangeable assistants. They are the components that carry decisions throughout the harness, each responsible for a distinct source of uncertainty, including task ambiguity, missing evidence, realization burden, temporal validity, and premature completion. This decomposition matters because next-generation time series problems rarely fail at only one point. A system may misunderstand the task specification, search too narrowly, overfit one convenient strategy, or finalize a result before temporal claims have really been checked. The agent group therefore separates orchestration, requirement recovery, evidence expansion, solution realization, and independent review into distinct roles, so each stage can be questioned by another rather than silently inherited by the next. To keep the remainder of the appendix consistent, we use one stable shorthand for these six roles. The global orchestration role is the \hl{\emph{Orchestrator}}, the task interpretation role is the \hl{\emph{Interpreter}}, the external evidence role is the \hl{\emph{Evidence Collector}}, the solution construction role is the \hl{\emph{Constructor}}, the time series review role is the \hl{\emph{Temporal Governor}}, and the minimal context final recheck role is the \hl{\emph{Final Reviewer}}.

\subsubsection{Global Orchestration and Task Integration (\hl{\emph{Orchestrator}})}\label{sec:appendix-agent-orchestrator}

This role sits at the top of the agent group and turns the full harness into a directed execution process rather than a loose accumulation of local actions. Its function is to preserve coherence across task understanding, evidence growth, branch choice, and review, so the overall system can behave as a governed workflow under explicit temporal constraints.

The global orchestration role acts as the executive center of the harness. Its job is not to solve the whole task alone, but to decide which specialized role should enter next, which evidence and constraints must be carried across steps, and whether the current branch is still worth pursuing. In \method, this role initializes context anchors, reads protocol constraints, dispatches downstream roles, and keeps review checkpoints active around important actions. At the paper level, it is the component that binds task interpretation, evidence expansion, solution construction, and independent review into one controlled loop rather than a loose sequence of prompts.

Time series tasks often expose several plausible strategies at once, such as direct forecasting, retrieval-aided reasoning, benchmark-specific heuristics, or external evidence expansion. The orchestration role is therefore designed to keep several promising branches alive long enough for initial validation, instead of collapsing too early into a single recommended strategy. The design of \method also makes clear that orchestration is not the same as completion authority. For blockers, rollback, continuation, and completion decisions, the fixed review order is \hl{\emph{Final Reviewer}} $>$ \hl{\emph{Temporal Governor}} $>$ \hl{\emph{Orchestrator}} $>$ other subagents. The \hl{\emph{Orchestrator}} manages dispatch and execution, but it and other subagents must not override, weaken, or summarize away critic blockers, rollback requirements, or completion decisions. This makes the harness closer to a governed search process than to a monolithic agent.

\subsubsection{Task Interpretation and Task Specification (\hl{\emph{Interpreter}})}\label{sec:appendix-agent-interpreter}

Before any meaningful execution can begin, the harness must first decide what problem is actually being posed by the available task environment and what kind of evidence would count as adequate support. This role therefore operates as the task interpretation interface between raw user intent and the structured task object that downstream components can safely execute.

The task interpretation role is responsible for turning a user request and a heterogeneous task environment into an executable task specification. In \method, it rereads available materials, extracts goals, input assets, constraints, evaluation standards, missing assumptions, and delivery expectations, and it can even rewrite the original question set when upstream requirements were underspecified. This is particularly important for time series because realistic tasks often arrive with mixed inputs, partial column visibility, irregular temporal scope, or incomplete statements of what counts as success.

This role also makes the first task-type decision for the temporal problem itself. It distinguishes whether the task is forecasting, anomaly analysis, event reasoning, classification, or a mixed form, and it asks whether visual inspection, data interface normalization, or public benchmark interpretation should happen before modeling. In harness terms, this role ensures that the task specification is fully clarified before execution begins. The system does not begin with optimization; it begins by recovering what the task actually is.

\subsubsection{External Evidence Expansion (\hl{\emph{Evidence Collector}})}\label{sec:appendix-agent-evidence-collector}

Once the task has been typed, the next source of uncertainty is whether the visible workspace already contains enough information to justify narrowing the solution space. The evidence expansion role addresses this problem by treating search, retrieval, and external comparison as a structured phase of the harness rather than as an ad hoc supplement to local reasoning.

The external evidence role exists because realistic time series work is rarely self-contained inside the currently visible table. In \method, this role conducts search across official documentation, recent methods, public leaderboards, benchmark discussions, public task conventions, and related author or organization links, while also checking whether the search space has become broad enough to justify contraction. The methodological point is that evidence expansion is treated as a primary process, not as an occasional side lookup.

If this responsibility were merged into the \hl{\emph{Orchestrator}} or the \hl{\emph{Constructor}}, search would tend to become opportunistic and narrow. By separating it, the harness can explicitly ask whether it has \emph{collected enough}, whether a stronger method family is still missing, whether strong public solutions reveal hidden evaluation quirks, and whether the problem should be reframed before further execution. This role therefore expands the available workspace into an evidence base that includes both local materials and external temporal knowledge.

\subsubsection{Solution Construction and Experimental Realization (\hl{\emph{Constructor}})}\label{sec:appendix-agent-constructor}

After the task specification and evidence base have been stabilized, the harness still needs a role that turns those intermediate decisions into executable artifacts, observable outputs, and testable results. This is the construction role of the system, which we refer to as the \hl{\emph{Constructor}}. This is the point at which candidate branches cease to be hypothetical and become concrete experiments, interfaces, and deliverables.

The solution construction role is the part of the harness that turns the current task specification and evidence state into runnable artifacts, analyses, experiments, and deliverables. In \method, it is expected to make the smallest complete intervention, stabilize interfaces before expanding methods, prefer direct use or light adaptation before expensive training, and verify claims with observable evidence. This makes it more than a coding assistant in the narrow sense. It is the realization component of the harness.

For time series tasks, this role is also required to respect temporal interfaces, leakage boundaries, experiment organization, and post-experiment diagnosis. In \method, this means producing not only a metric, but also structured outputs, plots, failure analysis, and hypothesis-oriented follow-up such as attribution or residual examination. At the paper level, this role operationalizes solution construction, where candidate branches become observable evidence rather than plausible text.

\subsubsection{Time Series Review Control (\hl{\emph{Temporal Governor}})}\label{sec:appendix-agent-temporal-governor}

Even a well-organized and evidence-rich workflow can still drift into weak temporal claims if no role is explicitly tasked with refusing shallow progress. The time series review role is introduced for exactly this reason. It keeps temporal validity, search completeness, and completion control under active review throughout the execution cycle.

The time series quality role is the main review mechanism of the harness. It is not a passive reviewer invoked only at the end, but a standing critic that inspects important steps before and after they run. In \method, it checks whether temporal rules were respected, whether method family coverage was wide enough, whether direct-use strategies were considered before training, whether search was broad enough, whether forecast outputs satisfy task requirements, and whether experiment evidence really entered the report. In practice, this role can block weak candidate solutions, request rollback, and delay completion until temporal concerns have been resolved, unless the \hl{\emph{Final Reviewer}} later decides otherwise.

This design encodes a specific research claim. Time series reliability is not guaranteed by generic correctness checks alone. Many failures are structurally temporal, such as illegal horizons, weak leakage control, narrow method coverage, or superficially good scores without sufficient attribution and failure slicing. By binding a dedicated review role to these checks, \method makes temporal quality an active control signal inside the harness rather than a retrospective comment after the main work is already over.

\subsubsection{Minimal Context Final Recheck (\hl{\emph{Final Reviewer}})}\label{sec:appendix-agent-final-reviewer}

The final reliability question is whether the current result still appears complete when stripped of the execution narrative that produced it. This role is therefore designed as an independent reviewer. It revisits the artifacts from outside the main context and checks whether completion still survives when earlier assumptions, explanations, and local momentum are no longer trusted.

The final role is a minimal context reviewer that inspects the current artifacts as if it had not participated in the earlier process. In \method, it rereads the task goal, the initial prompt anchor, the real artifacts on disk, and the required protocols and rules, then asks whether the task is complete without relying on the \hl{\emph{Orchestrator}}'s own narrative of completion. If a context snapshot is present, this role may use it only to check snapshot freshness or missed blockers, not as the source of truth for final closeout. It also reopens visual evidence directly when figures, tables, or report artifacts already exist. This role exists to counter one specific risk of agentic systems. Once a long execution story accumulates, earlier assumptions and local successes can bias later review toward premature completion.

For time series work, this fresh perspective is especially valuable because evidence may look convincing inside the active context while still failing a new reader's review on forecast validity, figure interpretation, artifact completeness, or alignment with the original task. \method therefore requires this role to reopen the main loop whenever it finds a blocker, gap, rollback point, or promising next action, and its completion decision takes precedence over earlier approvals. In methodological terms, this is the final barrier that prevents the harness from mistaking a coherent story for sufficient evidence.

\subsection{Skills}\label{sec:appendix-skills}

Skills play a different role from agents. If agents are the components that carry decisions in the harness, then skills are its \emph{reusable procedural abstractions}, compact capability units that can be called repeatedly across roles without forcing every agent to rediscover the same interface, safety check, diagnostic routine, or execution pattern from scratch. This matters especially for time series tasks because many of the most important operations are not ideas used only once, but repeated actions, such as stabilizing data entry, reading figure-rich documents, checking forecast legality, widening the search space, or organizing experiments into traceable outputs. The skill family therefore acts much like a function library for the harness, except that each function is itself a structured operational prior.

The repository contains 17 concrete skill files. The appendix keeps the same functional coverage but groups a few implementation-level skills together when they serve one paper-level purpose. For example, manual and automatic workspace startup are discussed together under context anchoring, and the planning-related skills are discussed together under executable planning. The titles below are therefore stable paper labels, not a claim that each subsection maps to exactly one file.

\subsubsection{Workspace Initialization and Context Anchoring}\label{sec:appendix-skills-workspace-initialization-and-context-anchoring}

Before the harness can reason well, it needs a stable starting state. This skill establishes that state by aligning project notes, runtime artifacts, and memory anchors, so later execution begins from explicit context rather than from implicit conversational drift.

In \method, this skill initializes the run trace and memory anchors, records the initial task baseline, summarizes the current goal, inputs, and gaps, and reminds the main flow to read runtime protocols and evaluation requirements when the task enters a multi-stage lane. At a system level, it creates the minimal persistent context that allows later stages to remain comparable across a long execution chain.

Time series tasks are often multi-stage and evidence-heavy, so early drift about task goals, metrics, or non-goals can silently affect later decisions. By forcing context anchoring at the start, the harness keeps temporal assumptions and evaluation intent visible from the first step onward.

\subsubsection{Input Safety Screening}\label{sec:appendix-skills-input-safety-screening}

The harness also assumes that incoming materials are not automatically trustworthy. This skill is the generic risk filter before reading external inputs, acting on available materials, or going online, so execution remains bounded even when documents, attachments, or code artifacts contain misleading or dangerous content.

In \method, this skill identifies the input surface, the action surface, the risk level, and the safest next path before the system proceeds. It is repeatedly reused because safety is treated as a recurrent check, not as a startup check that appears only once.

Although this skill is domain general, it is especially important in a time series harness because many realistic tasks require external files, benchmark pages, or mixed source evidence. Without this check, the system could widen its evidence base at the cost of losing execution boundaries.

\subsubsection{Document and Attachment Intake}\label{sec:appendix-skills-document-and-attachment-intake}

Many time series tasks do not arrive as clean tables alone. They may be accompanied by reports, policy documents, scans, benchmark descriptions, or figure-rich attachments. This skill is designed to turn such materials into usable evidence without confusing extraction with execution.

In \method, this skill reads PDFs and related attachments in a strictly nonexecuting way, separates text extraction from OCR inference, and exports figures or tables when visual analysis is needed. This expands the workspace as an evidence source while preserving input boundaries.

For time series tasks, attachments often encode the contextual rules behind the data, such as event definitions, business constraints, policy timing, or visual trend evidence. This skill makes those sources legible to the harness before they are folded into task understanding or review.
\subsubsection{Data Interface Unification}\label{sec:appendix-skills-data-interface-unification}

The next recurrent problem is that data rarely enters the harness through a single canonical format. This skill addresses that issue by normalizing heterogeneous upstream sources into one shared data interface before downstream analysis, plotting, or experiments are allowed to proceed.

In \method, this skill explicitly covers tables recovered from documents, spreadsheet-style files, databases, and programmatic data loaders as four primary entry types, and it requires them to converge into a common specification of timestamps, targets, features, entities, splits, and storage location. This is the interface-level counterpart of temporal grounding.

If the data interface remains unstable, all later time series reasoning becomes unstable as well. By isolating normalization in a reusable skill, the harness prevents raw source quirks from leaking directly into modeling, evaluation, or reporting.

\begin{painpointbox}
    \textbf{Unified data interface.} A normalized time-series interface should make the following fields explicit before downstream work begins.
    \begin{itemize}
        \item \textbf{Temporal identity}: timestamp column, timezone, frequency, calendar convention, and forecast cutoff.
        \item \textbf{Entity structure}: entity identifier, hierarchy, grouping level, panel membership, and aggregation rule.
        \item \textbf{Model surface}: target variable, covariates, exogenous signals, static attributes, and missing-value policy.
        \item \textbf{Evaluation boundary}: train / validation / hidden split, admissible history, horizon length, and leakage-sensitive transforms.
        \item \textbf{Artifact binding}: source files, normalized table location, data-interface version, and plots or summaries produced from the interface.
        \item \textbf{Public tooling surface}: common choices include \texttt{pandas}, \texttt{polars}, \texttt{pyarrow}, \texttt{xarray}, \texttt{duckdb}, \texttt{pandera}, and \texttt{great\_expectations}.
    \end{itemize}
\end{painpointbox}

\subsubsection{Time Series Analysis Prior}\label{sec:appendix-skills-time-series-analysis-prior}

This is the most explicit skill devoted to time series in the whole library. Its purpose is to give the harness one reusable analytic frame for recognizing domain type, inspecting temporal structure, forming candidate method families, and reopening the problem when the observed sequence suggests a different interpretation.

In \method, this skill insists on domain recognition, data interface stabilization, visual analysis, structural feature analysis, and task typing before method choice. It operationalizes the principle already stated in the main text, which is to understand the series first and decide how to optimize only afterward.

Because this skill can be invoked by execution, search, or review roles alike, it prevents time series reasoning from becoming trapped inside only one stage of the harness. Instead, temporal knowledge becomes a shared prior reused throughout task definition, realization, and critique.
\begin{painpointbox}
    \textbf{Analysis checklist.} The reusable time-series prior keeps the following checks visible before method commitment.
    \begin{itemize}
        \item \textbf{Domain recognition}: identify whether the sequence comes from retail, energy, finance, healthcare, sensors, operations logs, or another mechanism-bearing domain.
        \item \textbf{Visual analysis}: inspect trend, seasonality, changepoints, spikes, missing spans, regime shifts, and multi-entity heterogeneity.
        \item \textbf{Structural features}: record horizon length, history length, lag candidates, calendar effects, event effects, and exogenous availability.
        \item \textbf{Task typing}: distinguish forecasting, anomaly detection, event reasoning, classification, imputation, segmentation, or mixed objectives.
        \item \textbf{Reasoning split}: separate alignment, slice comparison, relative change, lag / response order, structural pattern, and interaction reasoning.
        \item \textbf{Method prior}: keep simple baselines, statistical models, classical ML, deep learning, and foundation-model methods comparable before pruning.
        \item \textbf{Public method examples}: compare seasonal naive and moving-average baselines with \texttt{statsmodels}, \texttt{prophet}, \texttt{sktime}, \texttt{darts}, the Nixtla stacks, \texttt{gluonts}, \texttt{pytorch-forecasting}, and recent pretrained methods such as Chronos, TimesFM, MOMENT, or PatchTST-style backbones when task fit justifies them.
    \end{itemize}
\end{painpointbox}

\subsubsection{Forecast Output Control}\label{sec:appendix-skills-forecast-output-control}

Forecasting errors are often not only modeling errors, but also task specification errors. This skill prevents the harness from accepting outputs that look numerically plausible yet remain structurally invalid under the task specification.

In \method, this skill forces horizon checks, column and format checks, numerical plausibility checks, and uncertainty decisions before a forecast or structured temporal answer can enter the final answer or saved results. This turns the forecast itself into a reviewable artifact.

This skill gives the harness a reusable way to enforce one of the most fragile parts of temporal work. The central question is not whether the answer sounds confident, but whether it is legal, well formed, and plausible under the observed series and task boundary.

\begin{painpointbox}
    \textbf{Forecast output checks.} A forecast or temporal output is reviewable only after the harness completes the following checks.
    \begin{itemize}
        \item \textbf{Horizon}: start date, end date, step count, rolling-window convention, and hidden-period boundary.
        \item \textbf{Required columns and format}: required identifiers, row count, timestamp alignment, column names, output order, and submission format.
        \item \textbf{Numerics}: missing predictions, invalid signs, impossible magnitudes, duplicate rows, and unstable postprocessing.
        \item \textbf{Evidence boundary}: whether features, normalization, lag construction, and external signals respect the cutoff.
        \item \textbf{Uncertainty}: whether point forecasts are sufficient or intervals, caveats, scenario ranges, or risk flags are required.
        \item \textbf{Validation tools}: use column and quality helpers such as \texttt{pandera}, \texttt{great\_expectations}, \texttt{pydantic}, \texttt{evidently}, or lightweight custom checks before a forecast is accepted.
    \end{itemize}
\end{painpointbox}

\subsubsection{Tool Prior and Analysis Toolkit} \label{sec:appendix-skills-tool-prior}

Not every reasoning step should reopen the entire software ecosystem from zero. This skill gives the harness a reusable prior over frequently used \texttt{Python} tool families, so it can cover the method space quickly before spending time on fresh validation of only the most relevant candidates.

In \method, this skill organizes tool families across numerical computing, statistics, forecasting, deep learning, explainability, diagnosis, and data engineering. Its role is not to freeze a leaderboard, but to reduce repeated search cost and to improve early category coverage.

Time series work often needs to move repeatedly between feature extraction, classical baselines, deep models, interpretability, and diagnosis tools. By making that ecosystem legible through one skill, the harness gains a stable analysis toolkit instead of rebuilding its tool horizon at every step.

\begin{painpointbox}
    \textbf{Tool families covered by the analysis toolkit.} Operationally, the current prior covers the following reusable Python tool families.
    \begin{itemize}
        \item \textbf{Numerical computing and data engineering}
              \begin{itemize}
                  \item Core arrays and tables: \texttt{numpy}, \texttt{pandas}, \texttt{scipy}, \texttt{polars}, \texttt{pyarrow}, \texttt{xarray}.
                  \item Scale and acceleration: \texttt{dask}, \texttt{numba}, \texttt{jax}, \texttt{joblib}, \texttt{bottleneck}.
                  \item Symbolic, formula, sparse, and unit support: \texttt{sympy}, \texttt{patsy}, \texttt{sparse}, \texttt{pint}.
              \end{itemize}
        \item \textbf{Statistics, econometrics, Bayesian, and causal analysis}
              \begin{itemize}
                  \item Classical statistical and temporal models: \texttt{statsmodels}, \texttt{pmdarima}, \texttt{arch}, \texttt{prophet}, \texttt{orbit}, \texttt{pykalman}.
                  \item Bayesian and statistical testing support: \texttt{pymc}, \texttt{arviz}, \texttt{cmdstanpy}, \texttt{bambi}, \texttt{pingouin}, \texttt{scikit-posthocs}.
                  \item Causal and panel modeling: \texttt{econml}, \texttt{dowhy}, \texttt{causalml}, \texttt{causallib}, \texttt{linearmodels}, \texttt{lifelines}.
              \end{itemize}
        \item \textbf{Machine learning, AutoML, and explanation}
              \begin{itemize}
                  \item Classical learners and boosted trees: \texttt{scikit-learn}, \texttt{xgboost}, \texttt{lightgbm}, \texttt{catboost}, \texttt{river}.
                  \item Search, tracking, and AutoML: \texttt{optuna}, \texttt{ray[tune]}, \texttt{mlflow}, \texttt{wandb}, \texttt{pycaret}, \texttt{flaml}, \texttt{auto-sklearn}.
                  \item Explanation interfaces: \texttt{shap}, \texttt{lime}, \texttt{interpret}, \texttt{eli5}, \texttt{alibi}, \texttt{dice-ml}.
              \end{itemize}
        \item \textbf{Time series learning and forecasting}
              \begin{itemize}
                  \item General toolkits: \texttt{sktime}, \texttt{aeon}, \texttt{darts}, \texttt{kats}, \texttt{merlion}, \texttt{timeseria}.
                  \item Forecasting stacks: \texttt{statsforecast}, \texttt{mlforecast}, \texttt{neuralforecast}, \texttt{hierarchicalforecast}, \texttt{utilsforecast}, \texttt{datasetsforecast}, \texttt{gluonts}, \texttt{pypots}, \texttt{autogluon.timeseries}, \texttt{pytorch-forecasting}.
                  \item Temporal features, motifs, shifts, and anomalies: \texttt{pyts}, \texttt{tslearn}, \texttt{stumpy}, \texttt{ruptures}, \texttt{adtk}, \texttt{tsfresh}, \texttt{tsfel}, \texttt{pycatch22}, \texttt{tsaug}.
              \end{itemize}
        \item \textbf{Deep learning, quality checks, storage, and diagnosis}
              \begin{itemize}
                  \item Deep learning and pretrained-model infrastructure: \texttt{torch}, \texttt{lightning}, \texttt{tensorflow}, \texttt{keras}, \texttt{transformers}, \texttt{accelerate}, \texttt{flax}, \texttt{haiku}, \texttt{optax}, \texttt{tensorboard}, \texttt{tsai}, \texttt{neuralprophet}.
                  \item Signal, drift, and data checks: \texttt{antropy}, \texttt{emd-signal}, \texttt{pywavelets}, \texttt{librosa}, \texttt{spectrum}, \texttt{statsmodels.tsa}, \texttt{feature-engine}, \texttt{featuretools}, \texttt{evidently}, \texttt{great\_expectations}, \texttt{pandera}, \texttt{dirty\_cat}.
                  \item Data entry, query, and visualization: \texttt{openpyxl}, \texttt{xlsxwriter}, \texttt{duckdb}, \texttt{sqlalchemy}, \texttt{sqlite3}, \texttt{psycopg}, \texttt{pymysql}, \texttt{connectorx}, \texttt{s3fs}, \texttt{fsspec}, \texttt{fastparquet}, \texttt{orjson}, \texttt{matplotlib}, \texttt{seaborn}, \texttt{plotly}, \texttt{yellowbrick}, \texttt{scikit-plot}.
              \end{itemize}
    \end{itemize}
\end{painpointbox}

\subsubsection{Search Space Expansion}\label{sec:search-space-expansion}

One of the main failure modes of long-running agent systems is premature convergence onto a locally convenient strategy. This skill family exists to counter that tendency by widening the candidate space before contraction, both conceptually and through additional evidence.

In \method, the branch expansion skill opens multiple fundamentally different branches, keeps promising alternatives alive in the same wave, and asks for explicit comparison points before merging branches. This transforms search from a single path narrative into a controlled branching process.

In time series settings, widening the search space also means checking whether nearby formulations, strong public solutions, or practical precedents imply a better framing of the task. This skill therefore supports knowledge-grounded reasoning by preventing the harness from mistaking early convenience for sufficient coverage.

\begin{painpointbox}
    \textbf{Expansion axes.} Before narrowing the search space, the harness keeps several search directions explicitly alive.
    \begin{itemize}
        \item \textbf{Direct problem search}: query the stated task, dataset, benchmark, metric, and platform constraints.
        \item \textbf{Decomposed search}: split the task into domain, signal type, target variable, horizon, evaluation rule, and method family.
        \item \textbf{Related formulation}: test nearby framings such as event-driven forecasting, anomaly triage, regime detection, ranking, or control support.
        \item \textbf{Heuristic rewrite}: search synonyms, reverse questions, failure modes, input-output rewrites, and target-function variants.
        \item \textbf{Public-solution absorption}: keep leaderboard writeups, public notebooks, issue discussions, and strong baselines as a separate evidence branch.
        \item \textbf{Method-family expansion}: descend from domain-specific methods to general time-series forecasting, then expand across statistical, machine-learning, deep-learning, probabilistic, and foundation-model families.
    \end{itemize}
\end{painpointbox}

\begin{painpointbox}
    \textbf{Example: expanding a Sichuan power forecasting query.} A concrete search wave should not repeat only one phrase such as \texttt{Sichuan power forecasting}. It should branch into progressively wider but still relevant evidence surfaces:
    \begin{itemize}
        \item \textbf{Domain and policy context}: search for Sichuan electricity demand, hydropower share, power-market reform, electricity price policy, demand response, industrial load, weather shocks, holidays, and grid dispatch constraints.
        \item \textbf{Power-forecasting methods}: search load forecasting, electricity demand forecasting, probabilistic load forecasting, short-term load forecasting, renewable-aware forecasting, weather-aware forecasting, and province-level grid forecasting.
        \item \textbf{General time-series methods}: expand to ARIMA / SARIMA, ETS, state-space models, Kalman filtering, Prophet-style trend and holiday models, gradient boosting, random forests, and quantile regression.
        \item \textbf{Deep and probabilistic models}: include TCN, N-BEATS, N-HiTS, DeepAR, TFT, Informer / Autoformer / FEDformer style transformers, PatchTST-style patch models, and distributional or conformal forecasting.
        \item \textbf{Foundation and benchmark methods}: check recent time-series foundation models and benchmark toolkits such as Chronos, TimesFM, MOMENT, Lag-Llama, TimeGPT-style services, GIFT-Eval, and other current TSFM comparisons.
        \item \textbf{Executable evidence}: look for official docs, papers, reproducible repositories, public notebooks, datasets, metric conventions, and failure analyses before narrowing to one strategy.
    \end{itemize}
\end{painpointbox}

\subsubsection{Long Horizon Reasoning and Dependency Tightening} \label{sec:appendix-skills-long-horizon}

After branches are opened, the harness still needs a reusable way to decide which dependencies matter first and which rollback points must remain visible. This skill performs that tightening step.

In \method, this skill decomposes a complex task into assumptions, dependencies, validation order, rollback points, and candidate merge conditions. It is therefore not only a thinking aid, but a mechanism for protecting the execution order of the harness.

Time series workflows often combine data alignment, external evidence, model choice, experiment loops, and output control. By making those dependencies explicit, this skill reduces the chance that the harness validates the wrong thing first or hides an upstream temporal error inside a downstream metric.

\subsubsection{Executable Planning}\label{sec:executable-planning}

The harness also needs a planning capability that is operational rather than rhetorical. These skills establish that requirement by converting context into a looped execution skeleton with owners, checkpoints, parallel branches, and rollback rules.

In \method, planning is not treated as a paragraph written once, but as a dynamically updated loop that can be synchronized with task state, branch identifiers, comparison points, and review checkpoints. The related context anchoring skill complements this by ensuring that execution begins from aligned context.

For time series work, many steps are not naturally serial. Search, data stabilization, public solution reading, and validation may all need to stay alive together. This planning stage preserves that parallel structure while still making restart and completion conditions explicit.

\subsubsection{Critical Rollback and Gap Review}  \label{sec:appendix-skills-critical-rollback-and-gap-review}

Even with planning and review roles in place, the flow can still become complacent, repetitive, or overly eager to declare completion. This skill exists as a reusable rollback loop that detects such moments and forces rollback when evidence quality has not caught up with narrative progress.

In \method, this skill is triggered when the system keeps tuning one strategy without new information, tries to finish without adequate validation, leaves figures uninterpreted, or mistakes a first answer for a finished answer. It marks the point at which the harness should pause forward execution and reopen the problem instead.

This skill is especially useful for time series because many weak solutions look superficially acceptable until error decomposition, figure reading, or task specification review reveals otherwise. By packaging rollback logic into a reusable skill, the harness keeps evidence completion active even outside formal critic roles.

\begin{painpointbox}
    \textbf{Rollback triggers.} The harness should reopen the loop when any of the following patterns appear.
    \begin{itemize}
        \item \textbf{One-strategy fixation}: repeated tuning of the same strategy without new evidence or branch comparison.
        \item \textbf{Unconsumed artifacts}: figures, logs, metrics, tables, or retrieved sources exist but are not used in the claim.
        \item \textbf{Weak validation}: the candidate solution has a plausible output but lacks local checks, slice analysis, baseline comparison, or constraint review.
        \item \textbf{Temporal ambiguity}: horizon, cutoff, lag construction, target alignment, or feature availability remains uncertain.
        \item \textbf{Premature completion}: the system tries to finish while blockers, reviewer objections, or promising next actions remain unresolved.
    \end{itemize}
\end{painpointbox}

\subsubsection{Formal Report Construction}\label{sec:formal-report-construction}

The harness does not treat reporting as a final formatting stage. This skill is responsible for turning experiment results, figures, statistical tests, and interpretability artifacts into a traceable report in which evidence is actually consumed rather than merely left unused.

In \method, this skill requires explicit binding among conclusions, tables, figures, artifact references, and appendix materials, and it forbids empty report language unsupported by real results. It also enforces the rule that figures in the body must be followed by actual analysis.

Since time series evidence is often visual, comparative, and diagnosis-heavy, good reporting is part of final review in the harness. This skill ensures that plots, forecast checks, and failure analysis remain part of the final argument rather than being left as side products.

\subsubsection{Public Artifact Evidence Retrieval}\label{sec:public-artifact-evidence-retrieval}

Some evidence questions cannot be settled by generic web search or documentation summaries. This skill addresses that gap by treating public repositories, model hubs, and dataset platforms as primary evidence surfaces.

In \method, this skill traces source materials, issue discussions, revision histories, model cards, dataset cards, and related author or organization links across public repositories and model platforms. It is the mechanism that lets the harness verify whether a capability really exists and how it is realized in practice.

Unlike the broader search space expansion skill, this one is narrower and more confirmatory. Its contribution is not to widen the conceptual branch set, but to ground candidate methods in concrete public evidence before they influence solution choice.

\subsubsection{Reusable Template Scaffolding}\label{sec:reusable-template-scaffolding}

Not every new capability should be improvised from scratch each time it reappears. This skill provides the lightweight scaffold for turning an emerging repeated pattern into a reusable capability description.

In \method, this skill creates a minimal template containing trigger conditions, workflow, boundaries, and output format, so a new reusable routine can be structured quickly before a heavier formal version is written.

This matters because harnesses evolve through repeated operational patterns. A template skill allows the system to capture those patterns cleanly before they are lost in single run traces.

\subsubsection{Capability Evolution and Abstraction Reuse}\label{sec:capability-evolution-and-abstraction-reuse}

Once repeated gaps become visible, the harness needs a structured way to decide whether they should remain local workarounds or be saved as new reusable structure. This skill governs that decision.

In \method, this skill asks whether an uncovered ability is truly recurrent, whether it fits better as a reusable skill or a new role, and how it should be linked back into the current relation map. This protects the harness from both missing useful abstractions and unnecessary role proliferation.

Time series systems often accumulate domain-specific or task-specific routines over time. By formalizing when those routines should become explicit abstractions, this skill supports controlled capability growth rather than ad hoc expansion.

\subsubsection{Experiment Infrastructure and Reproducibility}\label{sec:experiment-infrastructure-and-reproducibility}

The final skill in this family is the one that turns experimentation into a governed infrastructure rather than a pile of temporary scripts. It is the reproducibility infrastructure of the harness.

In \method, this skill standardizes experimental organization, run configuration, artifact output, metric logging, ablation structure, and result analysis around one common layout. It also requires condition repair before a strategy is declared infeasible, which makes experimental failure itself more informative.

For next-generation time series tasks, reproducibility is inseparable from reliability. Data interfaces, evaluation boundaries, plots, and comparative runs all need stable locations and traceable outputs. This skill provides that infrastructure, allowing experimental evidence to reenter review and reporting without ambiguity.

\begin{painpointbox}
    \textbf{Experiment artifact package.} A run becomes reviewable when it leaves behind a compact but complete evidence bundle.
    \begin{itemize}
        \item \textbf{Configuration}: dataset slice, split policy, horizon, random seed, method settings, and resource assumptions.
        \item \textbf{Execution record}: command, script, environment note, runtime outcome, failure mode, and rerun instruction.
        \item \textbf{Scored outputs}: predictions, metrics, validation report, leaderboard or local-score response, and baseline comparison.
        \item \textbf{Diagnosis artifacts}: plots, residual summaries, error slices, feature attribution, and representative failure cases.
        \item \textbf{Report binding}: the exact tables, figures, and claims that consume the experimental evidence.
        \item \textbf{Public analysis tools}: typical support includes \texttt{numpy}, \texttt{pandas}, \texttt{scikit-learn}, \texttt{statsmodels}, \texttt{lightgbm}, \texttt{xgboost}, \texttt{shap}, \texttt{matplotlib}, \texttt{seaborn}, \texttt{plotly}, \texttt{mlflow}, and \texttt{optuna}.
    \end{itemize}
\end{painpointbox}

\subsection{Rules}\label{sec:appendix-rules}

Rules are the constraint layer of the harness. If agents carry decisions and skills provide reusable operational abilities, then rules specify the admissible boundary within which those decisions and abilities may act. Their role is to prevent local plausibility from being mistaken for global validity. A candidate solution may look promising while leaking future information, a search process may feel broad while remaining narrow, and a report may sound complete without consuming its own evidence. In a time series setting, this layer matters because many serious failures do not arise from one obviously broken model, but from weak handling of temporal constraints, thin evidence chains, and premature completion. The following rules define how coordination, temporal constraints, experimentation, search, autonomy, and runtime control are stabilized across the whole harness.

\subsubsection{Core Coordination Principles}\label{sec:core-coordination-principles}

This rule group defines the operating doctrine of the whole system. It does not target one specific role or one specific task type. Instead, it specifies the common expectations that make the other components behave like one harness rather than like several independent workers.

In \method, the core rules explicitly reject surface success as sufficient proof. A local metric gain, a clean-looking output, the temporary absence of an error, or confident wording is not treated as evidence that a candidate solution is sound. The same rules place leakage prevention above all other goals, requiring the system to isolate and record any suspected contamination before quality claims are made. The emphasis is simple. Evidence comes first, optimism second.

The same rule set also requires that execution remain proportionate but complete. Basic tasks should stay light, while formal tasks may expand into reports, plots, appendices, and traceable artifacts. It prefers reading local material before external search, consuming experimental evidence inside reports once it exists, and using visual analysis as another evidence surface when structure may be easier to see than to describe. The updated rules also make the review path explicit, treat local checkpoints as a detailed replay mechanism, and include final housekeeping as part of delivery expectations. This gives the harness one shared expectation of what a closed loop actually means.

\subsubsection{Time Series Constraint Checks}\label{sec:appendix-rules-time-series-constraint-checks}

The next rule group is the domain-specific constraint layer for temporal tasks. Whereas the core rules govern the harness in general, these rules govern what must already be true before a time series result can be regarded as meaningful at all.

In \method, these rules require explicit task typing, explicit time representation, and leakage prevention before quality is discussed. Timestamps, frequency, entity granularity, forecast step, exogenous variables, and boundary-respecting splits must all be made clear. The same rules keep missing values, outliers, drift, shocks, and regime changes visible rather than allowing them to disappear into a generic preprocessing story. Temporal constraints are the first check, not a later refinement.

The time series rules also insist that metrics match the real task target and that a simple baseline remain visible. A complex method is therefore never allowed to compare only against itself, and a forecasting metric is not treated as interchangeable with an event, ranking, or interval objective. In paper terms, these rules keep temporal validity and temporal comparison anchored to the same task specification.

\begin{painpointbox}
    \textbf{Minimum temporal constraint surface.} Before a time-series claim is treated as meaningful, the harness records the following surface.
    \begin{itemize}
        \item \textbf{Task type}: forecasting, classification, detection, segmentation, anomaly analysis, representation learning, or a mixed form.
        \item \textbf{Time representation}: timestamp format, timezone, frequency, forecast step, entity granularity, and calendar assumptions.
        \item \textbf{Boundary checks}: split definition, cutoff date, normalization boundary, feature-building boundary, and target alignment.
        \item \textbf{Data conditions}: missingness, outliers, drift, shocks, interventions, regime changes, and exogenous-variable availability.
        \item \textbf{Comparison standard}: baseline visibility, metric-target match, horizon-specific evaluation, and nonforecast utility when relevant.
        \item \textbf{Common metrics}: use task-matched measures such as MAE, RMSE, RMSLE, MAPE / sMAPE, MASE, pinball loss, CRPS, AUROC, F1, or service-level utility rather than a generic score by default.
    \end{itemize}
\end{painpointbox}

\subsubsection{Experimental Evidence Checks}\label{sec:experimental-evidence-checks}

This rule group governs what counts as credible experimental evidence. It is the part of the harness that prevents experimentation from collapsing into a sequence of loosely related runs whose outputs cannot later support a real conclusion.

In \method, experiment rules require the smallest meaningful baseline first, explicit data splits and stop conditions before the experiment matrix is expanded, and a clear distinction among execution failure, realization failure, and decision failure. Results must remain reproducible, traceable, and comparable, with configuration, metrics, and failure information preserved rather than buried in narrative history. This turns experimentation into a governed evidence process rather than a collection of single trials.

The same rules also require iterative diagnosis once a metric exists. If outside information is thin, the flow should shift to local validation, slice experiments, or minimal reproduction rather than stalling in vague uncertainty. Once results exist, the loop is expected to continue through error analysis, visual inspection, and post-experiment hypothesis analysis, including attribution, residual structure, failure cases, or equivalent statistical support. For time series tasks, this matters because a single aggregate score can conceal the actual temporal failure mode.

\subsubsection{Search and Source Expansion Policy}\label{sec:search-and-source-expansion-policy}

Search is also constrained by explicit rules, because knowledge gathering is one of the easiest places for the harness to become narrow while still appearing active. This rule group therefore governs how evidence should be expanded before it is trusted.

In \method, external search begins only after local material is checked, and it does not stop after one query or one convenient hit. By default, it expands several search axes in parallel, including direct problem search, decomposed search, related search, heuristic rewrites, and trend-oriented discovery. If a public leaderboard or a strong public solution exists, it remains a separate solution-absorption branch rather than a background note. The point is to make the harness earn the right to narrow its search space.

The same policy imposes a clear source order, beginning with official documentation, then official repositories, original papers, strong public implementations, and only after that broader community material. It also treats repositories and model or data platforms as primary evidence when behavior, assets, or recent evolution must be verified. At a methodological level, this rule group controls not only \emph{how much} the harness searches, but also \emph{what kind} of evidence it is allowed to trust.

\subsubsection{Role Autonomy and Review Authority}\label{sec:role-autonomy-and-review-authority}

The harness also needs explicit rules about what its roles are allowed to do to the problem definition itself. Without that boundary, every subagent would become a passive executor of the \hl{\emph{Orchestrator}}'s first wording, and the review chain would gradually lose its independence.
In \method, analysis-oriented subagents are explicitly allowed to ask whether the \hl{\emph{Orchestrator}} posed the wrong question, to reorder priorities when a more upstream issue appears, and to request visual analysis or renewed entry when new structure is still unresolved. This does not encourage disorder; it prevents the whole harness from inheriting a narrow framing merely because it was stated first.

The same rules also make critique substantive rather than rhetorical. Review roles are expected to report blockers, maintain explicit response-and-recheck loops, and keep unresolved risks visible until they are actually answered. In practice, review decisions take precedence over routine execution decisions, so blockers, completion objections, and rollback requests cannot be dismissed by lower-level roles. This is what allows the harness to resist its own momentum.

\subsubsection{Execution Surface Conventions}\label{sec:appendix-rules-execution-surface-conventions}

Finally, the harness needs a rule set for the execution surface itself. These conventions are not merely interface preferences. They are the runtime protocols that make planning, rollback, evidence collection, and completion behave in a structured and repeatable way.

In \method, these conventions govern how execution begins, how related task specifications are read, how planning chains are formed, and how task records or current states are updated. Complex work is expected to pass through explicit branch expansion, dependency tightening, and planning before long execution begins, and state rollback must be represented directly rather than hidden in narrative summaries. This gives the harness an execution surface that can be traced and revised.

The same conventions also define how the system is allowed to approach termination. A pre-completion check must reopen branch search and reasoning before completion, repairable condition gaps should be fixed and retried before a strategy is declared infeasible, and any serious final critique should return the system to a full new loop. The same runtime structure also treats final housekeeping and review integrity as explicit parts of completion. At a methodological level, these conventions supply the procedural structure that keeps all earlier components from dissolving back into informal prompt interaction.

\subsection{Memory}\label{sec:appendix-memory}

Memory is the \emph{persistent layer} of the harness. If rules constrain behavior and protocols organize runtime flow, memory preserves the stable state that prevents long executions from drifting, repeating the same mistakes, or forgetting why a current branch exists in the first place. \method makes an explicit distinction here. Run-specific process history belongs to the trace, while reusable priors, structural relations, decision anchors, and completion records belong to persistent memory. This distinction matters especially for time series tasks, where progress often unfolds over many stages of task recovery, evidence expansion, experiment refinement, and final review. Without memory, the harness may remain locally active but globally unstable.

The repository memory template directory contains separate files for persistent memory, positive and negative priors, relation maps, prompt anchors, context snapshots, progress, features, decisions, TODO mapping, completion gates, directory notes, and trace seeding. The appendix groups these files by role in the harness; it does not rename the runtime files or change their responsibilities.

\subsubsection{Persistent Positive Priors}\label{sec:persistent-positive-priors}

The first memory type stores what has repeatedly worked well enough to deserve reuse. It is not a place for generic praise or temporary success, but for verified defaults, strengthening actions, and reliable strategies that continue to help later tasks.

In \method, positive memory records validated priors, useful strategies, recommended defaults, and reusable experience, always with scope and limits attached. This prevents a local optimum from being mistaken for a universal rule.

For a time series harness, such priors are especially valuable when certain patterns repeatedly improve quality, for example early data interface stabilization, baseline-first validation, or pre-stop strengthening actions. Storing these patterns explicitly gives later runs a stable starting advantage without pretending that they apply everywhere unchanged.

\subsubsection{Persistent Negative Priors}\label{sec:persistent-negative-priors}

The complementary memory type stores what has already failed in a way that should not be rediscovered from scratch. This includes not only failed strategies, but also false confidence patterns and structural failure modes.

In \method, negative memory captures failed paths, invalid assumptions, risk patterns, and no-go zones, together with the evidence of why they failed and where the resulting caution still applies. It is more than a failure list; it is a reusable boundary record.

For time series tasks, this is one of the main defenses against repeated leakage, shallow metric optimism, poor history choice, or early stop bias. By keeping these failures explicit, the harness turns past error into future prevention rather than into forgotten local failure.

\subsubsection{Project Memory Ledger}\label{sec:project-memory-ledger}

The project memory ledger defines the internal structure of the whole memory system. Without such a ledger, persistent memory can easily degrade into an undifferentiated notebook in which reusable priors, run-specific process notes, and runtime state are mixed together.

In \method, this ledger explains the roles of the separate memory records and enforces the distinction between reusable memory and the current trace. Positive and negative priors, relation maps, prompt anchors, progress records, and completion artifacts are all given explicit places instead of being merged by convenience.

This structure is important because a harness is not only a sequence of actions but also a sequence of remembered decisions and assumptions. When the storage scheme is explicit, later review can tell whether a given statement is a persistent prior, a current task fact, or only a transient observation.

\subsubsection{Initial Prompt Anchor}\label{sec:initial-prompt-anchor}

Among all memory artifacts, the initial prompt anchor plays the strongest role in preventing drift. It fixes the earliest task wording, goals, explicit metrics, and non-goals, and keeps constraints in one place before long execution begins.

In \method, this record is a persistent baseline containing the original prompt snapshot and the first task anchors, while later clarifications are added without overwriting the initial block. This ensures that the harness does not silently revise its own starting task specification after the work has already evolved.

This matters for time series tasks because long workflows often tempt the system to drift from the original target toward whatever currently looks easiest to optimize. By preserving the original goal and non-goals explicitly, the harness can later ask whether the final artifacts still solve the initial problem instead of merely solving a later local surrogate.

\subsubsection{Progress Ledger}\label{sec:progress-ledger}

The progress ledger records the present movement of the harness rather than its longer term priors. It keeps the current focus, what has been finished in the current round, and what the most justified next step is.

In \method, this ledger is intentionally small. It tracks current focus, completed items for the round, and suggested next actions with evidence. It is a compact runtime state rather than a verbose diary.

For long time series workflows, where the same task may pass through interpretation, search, realization, diagnosis, and review in different cycles, such a snapshot is essential to keep the active stage legible across roles. It prevents the system from confusing motion with progress.

\subsubsection{Feature Ledger}\label{sec:feature-ledger}

The feature ledger separates what has already been delivered from what is still only planned. This distinction is especially important in complex runs, where partial implementation can otherwise be mistaken for real completion.

In \method, this memory record explicitly stores delivered features with evidence and planned features with preconditions and blockers. It makes capability state visible rather than leaving it implicit in scattered artifacts.

In a time series harness, deliverables may include not only model outputs but also evaluation interfaces, plots, data adapters, or review artifacts. The feature ledger keeps these visible as separate states, so the system does not speak about planned support as if it were already available.

\subsubsection{Decision Ledger}\label{sec:decision-ledger}

Not every important choice should remain buried inside transient reasoning. The decision ledger captures the structural choices that shape the run, together with the cost of undoing them and the cases where a decision was deliberately deferred.

In \method, this record stores structural decisions, their reasons, and rollback cost, while also preserving deferred decisions with explicit revisit triggers. The design keeps the harness from treating unresolved choices as if they had already been settled.

For time series tasks, this is particularly useful when choices about task framing, baseline retention, experiment strategy, or reporting boundary affect many downstream artifacts at once. The decision ledger keeps those pivots inspectable and reversible.

\subsubsection{Relation Map}\label{sec:relation-map}

The relation map records how the moving parts of the harness connect to one another. It is the memory artifact that preserves structural topology rather than task content.

In \method, this map describes role boundaries, call chains, pre-completion review order, and the relation between roles and gap-filling skills, often with an explicit structural diagram. Its function is to make the current orchestration pattern visible instead of leaving it implicit in separate prompts.

As the harness grows, roles and skills can only remain legible if their relations are remembered explicitly. For time series work, where review, search, execution, and final critique all interact, the relation map preserves the architecture of cooperation itself.

\subsubsection{Task Trace Record} \label{sec:appendix-memory-task-trace-record}

The task trace record is the closest memory neighbor to a run log, but it still has a more structured function than raw logging. It records how a single task run was assessed, changed, verified, and handed forward.

In the template used by \method, the trace captures task goal, self-review, change scope, verification, and follow-up suggestions for each round, while clearly distinguishing itself from persistent memory. It is therefore the historical record of one run, not the store of reusable truths.

This distinction is important because time series work often requires several rounds of rollback, plot-guided reinterpretation, or evidence strengthening. The trace preserves that path-dependent history, while the rest of the memory system remains reserved for reusable structure.

\subsubsection{Context Snapshot}\label{sec:appendix-memory-context-snapshot}

Long runs may exceed the useful working context of a single interaction. The context snapshot is the memory record that keeps the next round grounded without forcing every role to reread the whole history.

In \method, this record stores the current phase, task anchors, active blockers, forbidden actions, decisions still in force, verified evidence, default next-round dispatch focus, and history that can be safely ignored. It is derived from the initial prompt, progress ledger, decision ledger, TODO alignment map, completion gate, and current blocker set. It is not free-form memory and does not replace those source records.

For time series work, this record is useful because long execution often accumulates many temporary plots, experiments, and branch notes. The snapshot keeps only the constraints and evidence that still matter for later dispatch, while preserving the rule that source artifacts and critic conclusions override the snapshot if they conflict.

\subsubsection{Task Plan Alignment Map} \label{sec:appendix-memory-task-plan-alignment-map}

Planning only helps if the running task state can still be mapped back to it. This memory artifact preserves that map by aligning plan steps, action items, branch identifiers, frontier state, blockers, and rollback rules.

In \method, the map explicitly ties each plan step to an active action item, owner, wave, comparison point, update trigger, and rollback rule. It prevents execution from drifting away from the plan while still pretending to follow it.

For a harness that deliberately keeps multiple branches alive, this alignment memory is essential. It keeps branch structure, comparison points, and completion consequences visible even when the active flow has already moved several steps ahead.

\subsubsection{Completion Gate Record}\label{sec:completion-gate-record}

The completion gate record is the memory artifact closest to final review. Here, \emph{completion gate} refers to the implementation-level completion checklist: it records the checkpoints that must be cleared before the harness is allowed to treat the current task as complete.

In \method, this record tracks the pre-stop checks from branch expansion, deep reasoning, critic review, \hl{\emph{Temporal Governor}} review, and \hl{\emph{Final Reviewer}} review. It also records drift against the initial prompt, context-snapshot freshness, workspace cleanup, role challenges, remaining action count, completion state, stop permission, and the reason for that decision. This makes completion criteria explicit and reviewable.

This is especially important for time series tasks, where completion is often threatened by subtle remaining actions rather than obvious unfinished work. By storing the checkpoint status explicitly, the harness can show not only that it finished, but why completion was allowed.

\begin{painpointbox}
    \textbf{Completion gate fields.} A completion record is expected to preserve the state of the following template fields.
    \begin{itemize}
        \item \textbf{Pre-stop checks}: brainstorm, deep-reasoning, critic-loop, \hl{\emph{Temporal Governor}}, and \hl{\emph{Final Reviewer}} status.
        \item \textbf{Task alignment}: initial-prompt review and any drift from the original prompt.
        \item \textbf{Context freshness}: latest context-snapshot refresh trigger and stale-risk status.
        \item \textbf{Role challenges}: whether any role can still continue and what next action remains.
        \item \textbf{Completion decision}: remaining action count, completion state, stop permission, and reason.
        \item \textbf{Housekeeping}: workspace cleanup status, removed leftovers, and reasons for keeping remaining artifacts.
    \end{itemize}
\end{painpointbox}

\subsubsection{Artifact Boundary Reference}\label{sec:artifact-boundary-reference}

The final memory artifact preserves the boundary between different record types. This may look mundane at first glance, but in a multicomponent harness it is one of the practical defenses against confusing one artifact with another.

In \method, this reference explains how runtime trace differs from persistent memory, how version history differs from abstract memory, and how agents, protocols, evaluations, skills, and memory records should be interpreted as distinct record types. It also makes clear that local checkpoints are only a detailed replay mechanism, not a substitute for memory or trace. This makes artifact boundaries explicit rather than leaving them to informal interpretation.

A harness that accumulates many artifact types can remain stable only if those boundaries continue to mean the same thing over time. For time series work, where data interfaces, plots, evaluations, reports, and runtime records all accumulate together, artifact boundaries are part of operational memory, not an afterthought.

\subsection{Evaluation}\label{sec:appendix-evaluation}

Evaluation is the measurement and acceptance layer of the harness. If memory preserves state and rules constrain behavior, evaluation decides how the current run should actually be measured, compared, aggregated, and accepted. This layer is therefore not a single scalar metric attached at the end of execution. It is a set of evaluation records that ranges from suite design to run-level grading, aggregation across runs, regression coverage, and final release readiness. For time series tasks, this matters because many meaningful failures do not appear as one bad number alone. Horizon illegality, protocol violations, thin robustness, or weak holdout handling may all remain hidden unless the evaluation layer is explicit.

\subsubsection{Evaluation Suite Design} \label{sec:appendix-evaluation-suite-design}

The suite is the highest level evaluation object in this stack. It defines what class of tasks should be tested together, which metrics matter, how tasks are selected, and what kind of iteration or validation rhythm the harness is expected to follow.

In \method, a suite specification includes the goal, metric requirements, task types, task selection policy, local validation policy, platform prior policy, error analysis policy, modeling reflection policy, iteration policy, trial policy, and required artifacts. The suite does not merely enumerate tasks; it formalizes the evaluation regime under which they become comparable.

This design is especially important for open-ended time series work, where evaluation must often say not only what the main metric is, but also when to switch from outside search to local validation, how to handle benchmark-specific heuristics, and how error analysis should continue after the first score appears. The suite therefore defines the shape of valid evaluation before any individual run is graded.

\subsubsection{Grading Interface}\label{sec:grading-interface}

Once the suite defines the set and the policy, the next question is how one concrete run is actually graded. The grading interface answers that question by mapping the produced artifacts of a run into explicit metrics and verdicts.

In \method, a grader declares its input surface, output surface, grading logic, and the decision rule by which process records or outcome artifacts are turned into a metric and a verdict. It must also state how, if at all, a reference solution enters the grading process.

This interface is what prevents the harness from relying on vague assessments such as \texttt{looks correct} or \texttt{seems strong enough}. For time series work, where structured outputs, forecast validity, and protocol compliance may all matter at once, the grading interface turns one run into an auditable evaluation record instead of a subjective impression.

\subsubsection{Score Aggregation Across Runs}\label{sec:score-aggregation-across-runs}

Run-level verdicts are not yet enough to characterize a system. The aggregation stage is responsible for lifting those verdicts across multiple trials, tasks, and grading dimensions so that overall behavior can be discussed in a stable and comparable way.

In \method, the scorecard specification requires explicit metrics, an uncertainty policy, an aggregation rule, and report sections that include not only point estimates but also uncertainty and failure modes. It assumes from the start that the outcome of a harness cannot be reduced to one accuracy-like number.

For time series tasks, this is particularly valuable because variability across runs, horizons, or evaluation slices can matter as much as a central score. A scorecard defined across runs makes room for robustness, cost, latency, protocol compliance, and later time series criteria such as leakage or horizon validity, so system quality is not reduced to one number.

\subsubsection{Regression Comparison Matrix}\label{sec:appendix-evaluation-regression-comparison-matrix}

The regression matrix complements the suite by tracking what must not be lost as the harness evolves. Instead of asking only whether a new strategy works, it asks whether old capabilities and known risk areas remain covered after the change.

In \method, this matrix is defined by capability axes, risk axes, required coverage, and a holdout policy. Its purpose is to make the shape of regression protection explicit rather than hiding it behind a generic phrase like \texttt{regression tests exist}.

This matters because a harness may improve one dimension while silently damaging another. In time series settings, the lost dimension might be temporal validity, protocol compliance, or robustness under certain history or event conditions. The regression matrix gives those axes a stable comparison surface, so gains do not erase previously working behavior.

\subsubsection{Release Readiness Gates}\label{sec:appendix-evaluation-release-readiness-gates}

The final evaluation artifact is the release gate. Here, \emph{release gate} refers to the implementation-level release-readiness template: it translates the outputs of suites, graders, and scorecards into an explicit decision about whether progression is allowed, whether caution is required, or whether rollback must occur.

In \method, release gates separate hard thresholds, warning thresholds, rollback conditions, and manual review conditions. This means that completion is not reduced to a single \texttt{pass/fail} switch. Instead, the release-gate record preserves the structure of evidence needed for acceptance, caution, or return.

For a time series harness, this final stage is what keeps completion aligned with real quality rather than local momentum. A candidate solution may show promise but still fail a hard threshold, trigger a rollback condition, or require manual review because its risk profile is not yet acceptable. The release gate is the evaluation-side counterpart of final review in the rest of the harness.

\subsection{Protocols}\label{sec:appendix-protocols}

Protocols are the runtime interface layer of the harness. If rules define behavioral boundaries, memory preserves stable state, and evaluation formalizes measurement, then protocols specify how work, state, critique, and control signals actually move from one component to another during execution. Their value lies in reducing ambiguity. Without protocols, role handoff, blocker response, completion checks, and memory updates would all collapse back into free-form natural language. For a time series harness, this matters because many important failures are not local modeling failures, but failures of coordination, review transfer, or premature completion. Protocols therefore give the harness a structured execution interface rather than a loose conversational flow.

\subsubsection{Dispatch Interface}\label{sec:appendix-protocols-dispatch-interface}

The first runtime problem is how one role hands work to another without silently dropping assumptions, rights, or constraints. The dispatch interface addresses that problem by turning role handoff into an explicit packet rather than a vague instruction.

In \method, the dispatch packet includes the objective, known inputs, context mode, explicit focus, required reads, permitted action scope, constraints, autonomy settings, output requirements, and blocking context. This means a dispatched role does not merely receive a topic. It receives a bounded operating surface.

This is especially important in a time series harness because different roles may need different slices of context. The \hl{\emph{Final Reviewer}} should receive minimal context, while the \hl{\emph{Constructor}} may need broader evidence and a wider action scope. The dispatch protocol preserves those differences explicitly, which prevents role confusion from being hidden inside prose.

\subsubsection{Lifecycle State Model}\label{sec:appendix-protocols-lifecycle-state-model}

Once work is handed off, the next question is how the system knows where that work currently stands. The lifecycle protocol answers this by giving every subagent one shared state model instead of allowing each role to invent its own informal progress language.

In \method, the lifecycle model requires states such as assignment, context alignment, input validation, execution, self-review, report preparation, completion, and blockage, together with transition reasons and expected next states. The protocol therefore makes progress state explicit and comparable across roles.

For a long-running time series task, this matters because completion should not mean only that some local output exists. It should mean that context was aligned, input was validated, execution happened, and self-review was completed. The lifecycle protocol prevents those stages from disappearing into an unstructured summary.

\subsubsection{Reportback Interface}\label{sec:reportback-interface}

If dispatch structures the beginning of a work slice, reportback structures its return. This protocol is what allows the \hl{\emph{Orchestrator}} to integrate role outputs through stable fields rather than by guessing from writing style or confidence level.

In \method, every reportback must specify status, completed work, remaining gaps, new risks, suggested memory updates, possible rule or protocol updates, the suggested next step, the suggested next role, follow-up actions, whether the role can continue by itself, self-critique, and why stopping is not yet allowed. The same structure applies to the \hl{\emph{Final Reviewer}}, and lower components are not allowed to soften critic blockers or completion conclusions in summary. This turns role output into an interface object rather than a prose paragraph.

For time series work, where evidence may remain partial even after meaningful progress, a structured return is essential. It allows the harness to distinguish between a locally finished slice and a globally closable state, and it preserves the remaining gaps that must still be carried into later review.

\subsubsection{Review Response and Recheck Interface} \label{sec:appendix-protocols-review-response-and-recheck-interface}

The harness also needs a formal way to respond when review blocks a candidate solution. The review-response protocol provides that mechanism by turning \texttt{please address this issue} into an issue-by-issue exchange rather than a loose promise to improve.

In \method, a blocked role must answer each issue with acceptance status, current evidence, fix plan, missing evidence, remaining blocker, and request for recheck, while the critic must return a structured recheck verdict with reason, remaining gap, required next change, completion signal, and follow-up action. This makes blocker resolution a loop with explicit fields rather than a one-shot comment.

This protocol is particularly important for time series review because many blockers involve subtle questions of temporal constraints, evidence sufficiency, or output plausibility. A structured response loop prevents these issues from being softened or lost between review rounds.

\subsubsection{Completion Control Interface}\label{sec:completion-control-interface}

The next protocol governs the most consequential control signal in the whole harness, namely whether execution should continue, return for recheck, or is genuinely complete. It exists to keep completion signals stable across the system.

In \method, the protocol formalizes only three signals, \emph{continue}, \emph{rebuttal-required}, and \emph{allow-stop}. Each signal carries explicit reasons, scope, next action, completion checks, remaining action count, and complete-state status. In particular, \emph{allow-stop} is defined as a fully complete state, not as a soft phrase meaning close enough. When review signals conflict, the final completion decision stays with the \hl{\emph{Final Reviewer}}.

For time series tasks, this semantic precision matters because premature completion often arises from ambiguity rather than from explicit error. By requiring all completion checks to pass, including search coverage, figure analysis, and visual retest logic, the completion-control protocol prevents the harness from mistaking narrative exhaustion for evidence completion.

\begin{painpointbox}
    \textbf{Completion signal semantics.} The runtime interface allows only three completion-related signals.
    \begin{itemize}
        \item \textbf{Continue}: evidence is incomplete, an action remains useful, or a required check has not yet been reached.
        \item \textbf{Rebuttal-required}: a reviewer has raised a blocker that must be answered issue-by-issue before the candidate solution can proceed.
        \item \textbf{Allow-stop}: the task is complete, remaining action count is zero, completion checks are satisfied, and higher-priority reviewers do not object.
        \item \textbf{Reason field}: every signal carries scope, justification, next action if any, and the specific check or reviewer that produced it.
        \item \textbf{Authority rule}: final completion decisions are not weakened by lower-level summaries; stricter reviewer signals remain binding until resolved.
    \end{itemize}
\end{painpointbox}

\subsubsection{Runtime Event Logging} \label{sec:appendix-protocols-runtime-event-logging}

Beyond stable artifacts and stable signals, the harness also needs a record of key state changes as they occur. The runtime event protocol provides that record by encoding important transitions as structured events.

In \method, events such as dispatch creation, reportback receipt, rebuttal opening, rebuttal review, stop-go signal emission, completion gate update, context compaction, artifact synchronization, repository initialization, and local checkpoint creation are written with event type, source, task identifier, timestamp, summary, and affected artifacts. This does not replace the artifacts themselves. It records the movement among them.

In a long multirole time series workflow, many important shifts are process shifts rather than content changes. An event stream makes those shifts replayable and checkable, which is especially useful when later review needs to understand how a completion decision, blocker loop, or artifact update actually emerged.

\subsubsection{Memory Synchronization Interface} \label{sec:appendix-protocols-memory-synchronization-interface}

Finally, the harness needs a controlled way for subagents to influence persistent state without directly rewriting global memory. The memory synchronization protocol provides that write-back boundary.

In \method, subagents may only propose updates through the fields \texttt{positive\_candidate}, \texttt{negative\_candidate}, \texttt{relation\_candidate}, \texttt{decision\_candidate}, \texttt{context\_snapshot\_candidate}, \texttt{trace\_event\_candidate}, and \texttt{runtime\_event\_candidate}, together with explicit confidence and reason. The \hl{\emph{Orchestrator}} then decides whether to absorb those suggestions after reading the current artifacts first. Direct blind overwrite from a subagent suggestion is forbidden.

This protocol protects the memory system from uncontrolled mutation while still allowing useful lessons from local work to be preserved. For time series tasks, where repeated priors and repeated failure modes are especially valuable, such a controlled write-back path is what turns local findings into stable harness knowledge without sacrificing auditability.

\subsubsection{Context Compaction Interface}\label{sec:appendix-protocols-context-compaction-interface}

The protocol set also includes a context compaction interface for long-running sessions. Its job is to define when the main agent refreshes the context snapshot and when a dispatched role should receive compacted context rather than full history.

In \method, the snapshot must be refreshed after workspace initialization, after a major plan or strategy switch, after parallel reportbacks are merged, after a review-response loop opens, and before the \hl{\emph{Final Reviewer}} runs. For non-final-review roles in later rounds, dispatch normally points to the compacted context snapshot; a full-context dispatch must state why the snapshot and supporting artifacts are insufficient.

This protocol matters for time series tasks because a long run can accumulate a large amount of local history that is no longer relevant. Compaction keeps active blockers, still-valid decisions, verified evidence, and next dispatch focus visible, while making clear that the snapshot cannot override source artifacts or critic conclusions.

\section{Governed Execution and Delivery}\label{sec:appendix-governed-execution}

\subsection{Overall Flow}\label{sec:overall-flow}

The previous sections describe the harness as a set of components. This section explains how those components are activated together in one real run. In \method, execution does not proceed as a single forward chain from prompt to answer. It proceeds as a governed loop in which task grounding, branch expansion, temporal validation, rechecking, and final delivery are all allowed to reopen one another when evidence is still weak. This loop structure is especially important for time series tasks, because the main risk is often not that the system cannot produce an answer, but that it can produce one too early under weak temporal constraints.

At a high level, the cycle has three phases. The first phase stabilizes the task and expands the candidate space. The second phase turns candidate branches into executable evidence under temporal and evaluation checks. The third phase decides whether the current state is actually deliverable, or whether the flow must return upward through critique and rollback. What makes the whole process specifically a time series harness is that temporal grounding, temporal constraints, and temporal review are not confined to one stage. They remain active from the first interpretation of the task until the final completion decision.

\subsection{Entry and Expansion}\label{sec:entry-and-expansion}

The first phase of the governed loop is to make the problem legible before it becomes executable. The harness begins by anchoring the run, recovering the true task specification, stabilizing data entry, and only then widening the candidate search space. This order matters because branch expansion is only useful when the system already knows what kind of temporal object it is working with and which boundary conditions must remain intact.

\subsubsection{Entry and Task Grounding}\label{sec:entry-and-task-grounding}

The run begins by creating a stable starting state rather than by launching directly into method choice. The harness first anchors the initial prompt, initializes runtime artifacts, rereads the current workspace, and recovers the executable task specification from the user request and the visible evidence. Only after that does it normalize the data entry and expose the relevant temporal object for analysis.

In practical terms, the opening sequence binds together context anchoring, task interpretation, and data interface unification. The system asks what the real target is, which artifacts belong to the current task, where the temporal boundary lies, and what output will count as valid. In a time series setting, this stage also includes plot-first grounding, so the series can be inspected visually before the harness commits to a modeling strategy.

If the run starts from an unstable task definition, later search and optimization will only become more expensive versions of the wrong problem. For time series work, the danger is especially acute because horizon scope, leakage boundaries, and data alignment are often implicit. Entry and grounding therefore serve as the first barrier against drift in the whole system.

\subsubsection{Branch Expansion and Evidence Growth}\label{sec:branch-expansion-and-evidence-growth}

Once the task is grounded, the next priority is to avoid premature convergence. The harness enters an expansion phase in which multiple promising branches are kept alive in parallel long enough to produce initial information gain. Search in this phase is not a background convenience. It is one of the main mechanisms by which the harness prevents itself from mistaking local familiarity for global adequacy.

When public benchmark structure exists, this phase keeps at least two major lines visible. One line is self-directed exploration, and the other is public solution absorption. More generally, the harness uses a breadth-first wavefront logic, allowing promising branches to survive within the same wave until the first comparison point is actually reached. External evidence growth, method family widening, and related search expansion all belong to this phase.

For time series tasks, widening the branch set is especially important because many tasks are ambiguous between several candidate hypotheses at the start. The problem may turn out to be best handled as forecasting, event-conditioned reasoning, anomaly analysis, or a hybrid strategy. By expanding evidence and branches before collapse, the harness gives itself room to discover which temporal framing is actually informative.

\subsection{Execution and Re-entry}\label{sec:execution-and-re-entry}

The second phase is where candidate branches become concrete. However, execution is not treated as a free zone in which the chosen branch simply runs until completion. It remains bounded by temporal constraints, experiment checks, and evaluation checks, and any blocker raised during this phase may still force the flow back into an earlier stage of search or task reconstruction.

\subsubsection{Execution Under Temporal and Evaluation Checks}\label{sec:execution-under-temporal-and-evaluation-checks}

Once a branch is selected for concrete work, the harness turns it into analyses, experiments, and artifacts under explicit checks. The \hl{\emph{Constructor}} does not simply pursue whatever seems promising. It moves forward only while temporal constraints, forecast horizon, experiment protocol, and evaluation interfaces remain aligned with the original task.

In this phase, the \hl{\emph{Constructor}}'s work, local validation, artifact generation, plotting, and structured result writing are all treated as part of one evidence chain. Local validation is preferred before scarce external submission opportunities, and outputs are expected to remain compatible with the grading and score aggregation records that will later consume them. The construction phase is therefore not only productive; it is already aligned with evaluation.

Time series execution can fail in ways that look superficially productive. A model may train, a figure may render, and a forecast may have the right shape even when forecast horizon, leakage boundaries, or column and format validity are already broken. By keeping time series checks and evaluation interfaces active during execution itself, the harness prevents downstream review from becoming the first place where these errors are noticed.

\subsubsection{Rollback, Recheck, and Loop Re-entry}\label{sec:rollback-recheck-and-loop-re-entry}

The core reason this harness remains stable is that failure is not treated as a minor local defect to be patched quietly. It is handled as a return signal. When blockers appear, the system enters recheck mode, answers them point by point, and if the blockers still stand, it reenters the main loop from a more upstream stage rather than pretending that the candidate solution can still close cleanly.

At the procedural level, this phase is driven by review before and after key steps, the recheck protocol, critic escalation, and explicit rollback to earlier stages such as requirements recovery, renewed branch expansion, or deeper reasoning. A blocker is therefore not just a comment. It is a control event with consequences for task state and candidate-solution legitimacy.

For time series tasks, the most important blockers are often structural rather than superficial, including weak evidence, poor method family coverage, unstable history choice, or illegal forecast shape. By treating these issues as reasons to reopen the loop rather than to patch the current answer at the surface, the harness can still correct course even after substantial work has already been invested in one branch.

\subsection{Completion and Delivery}\label{sec:closeout-and-delivery}

The final phase of the loop is not merely a shutdown step. It is a completion process in which the harness asks whether the current state is genuinely deliverable under all of its own checks. The full review stack becomes most visible here. Branch search must be exhausted, reasoning must confirm that no promising path remains, temporal review must approve completion, and the \hl{\emph{Final Reviewer}} must still agree after rereading the artifacts from outside the main narrative.

In practice, this phase passes through an explicit pre-completion sequence of branch widening, dependency review, temporal review, and \hl{\emph{Final Reviewer}} inspection, together with completion gate checks, artifact existence checks, report evidence consumption, and final housekeeping. The key point is that review still controls completion: the \hl{\emph{Orchestrator}} may organize delivery, but final completion decisions remain on the review side of the harness.

Only after those checks pass does the harness treat the current state as deliverable. At that point, the system is allowed to bind real artifacts, a cleaned final state, and validated conclusions into final delivery. For time series work, this completion process distinguishes a reliable harness from a merely productive one. The system is assessed by whether it can justify ending under temporal, evidential, and review pressure, not only by whether it can end.